\documentclass[utf8]{FrontiersinHarvard} % for articles in journals using the Harvard Referencing Style (Author-Date), for Frontiers Reference Styles by Journal: https://zendesk.frontiersin.org/hc/en-us/articles/360017860337-Frontiers-Reference-Styles-by-Journal
\usepackage{url,hyperref,lineno,microtype,graphicx}
\usepackage[onehalfspacing]{setspace}

\def\keyFont{\fontsize{8}{11}\helveticabold }
\def\firstAuthorLast{McDonald-Bowyer {et~al.}} %use et al only if is more than 1 author
\def\Authors{A.  McDonald-Bowyer\,$^{1,*}$, S. Dietsch\,$^{1*}$, E. Dimitrakakis$^{1}$, J. M. Coote$^{1}$, L. Lindenroth$^{1}$, D. Stoyanov$^{1}$ and A. Stilli\,$^{1}$}
\def\Address{$^{1}$Welcome/EPSRC Centre for Interventional and Surgical Sciences, University College London, London, W1Q 7EJ, U.K}
\def\corrAuthor{Agostino Stili}

\def\corrEmail{a.stilli@ucl.ac.uk}

\begin{document}
\onecolumn
\firstpage{1}

\title[Organ Shape Sensing Using PAF Rails]{Organ Curvature Sensing Using Pneumatically Attachable Flexible Rails in Robotic-Assisted Laparoscopic Surgery} 

\author[\firstAuthorLast ]{\Authors} %This field will be automatically populated
\address{} %This field will be automatically populated
\correspondance{} %This field will be automatically populated

\extraAuth{}% If there are more than 1 corresponding author, comment this line and uncomment the next one.
% \extraAuth{corresponding Author2 \\ Laboratory X2, Institute X2, Department X2, Organization X2, Street X2, City X2 , State XX2 (only USA, Canada and Australia), Zip Code2, X2 Country X2, email2@uni2.edu}

\maketitle
\begin{abstract}

%%% Leave the Abstract empty if your article does not require one, please see the Summary Table for full details.
\section{}
In robotic-assisted partial nephrectomy, surgeons remove a part of a kidney often due to the presence of a mass. A drop-in ultrasound probe paired to a surgical robot is deployed to execute multiple swipes over the kidney surface to localise the mass and define the margins of resection. This sub-task is challenging and must be performed by a highly-skilled surgeon. Automating this sub-task may reduce cognitive load for the surgeon and improve patient outcomes. 
The eventual goal of this work is to autonomously move the ultrasound probe on the surface of the kidney taking advantage of the use of the Pneumatically Attachable Flexible (PAF) rail system, a soft robotic device used for organ scanning and repositioning. First, we integrate a shape-sensing optical fibre into the PAF rail system to evaluate the curvature of target organs in robotic-assisted laparoscopic surgery. Then, we investigate the impact of the PAF rail’s material stiffness on the curvature sensing accuracy, considering that soft targets are present in the surgical field. Finally, we use shape sensing to plan the trajectory of the da Vinci surgical robot paired with a drop-in ultrasound probe and autonomously generate an Ultrasound scan of a kidney phantom.

\tiny
 \keyFont{ \section{Keywords:} Medical robotics, soft robot applications, soft robot materials and design, soft sensors and actuators, shape sensing, da Vinci Research Kit, surgical robotics, robotic-assisted surgery} %All article types: you may provide up to 8 keywords; at least 5 are mandatory.
\end{abstract}

\section{Introduction}

Partial nephrectomy is a laparoscopic surgical procedure in which a portion of the kidney is removed. This operation maximises the patient’s postoperative kidney function compared to total nephrectomy\cite{Kaul2007DaFollow-Up} as it preserves renal function.
Robotic-Assisted Partial Nephrectomy (RAPN) employs robotics during this complex procedure and improves patient outcomes, as detailed in \cite{Kaul2007DaFollow-Up} and \cite{Bhayani2008DaTechnique}. Namely, it shortens hospital stays, reduces post-operative pain, and minimizes recovery time. Robotic assistance for high precision surgical tasks can also reduce surgeon fatigue \cite{Stefanidis2010RoboticWorkload} \cite{VanDerSchatteOlivier2009ErgonomicsSurgery}, thus improving accuracy. The RAPN surgical procedure is described in detail in \cite{Bhayani2008DaTechnique}. 
Tumour margin identification can be done in preoperative imaging modalities, Computer Tomography  \cite{Su2009AugmentedRegistration} and Magnetic Resonance Imaging  \cite{Shingleton2001PercutaneousGuidance}, and this information can be visualised intraoperatively through 3D model visualisation and image-guided navigation \cite{Ferguson2018TowardRe-registration}. Practically however, most RAPN surgeons utilize intraoperative ultrasound (US) to  evaluate tumour margins during the procedure \cite{Hekman2018IntraoperativeNephrectomy}. 
In \cite{Kaczmarek2013ComparisonNephrectomy}, the authors suggest that robotic and laparoscopic approaches have comparable perioperative outcomes when they help surgeons to guide drop-in US probes during the kidney scan, with the former having the advantage of increased surgeon dexterity and potential autonomy.

Despite the additional degrees of freedom (DoF) provided by robotic laparoscopic tools over hand-held laparoscopic tools, RAPN is still a challenging procedure that requires a highly skilled surgeon with years of speciality training. US scanning of the kidney is also a demanding sub-task due to slippage of the US probe on the organ surface, and needing to maintain contact between the tissue and the probe while also doing image interpretation with respect to tool actions \cite{Kaczmarek2013RoboticOutcomes}.
\begin{figure*}
    \centering
    \includegraphics{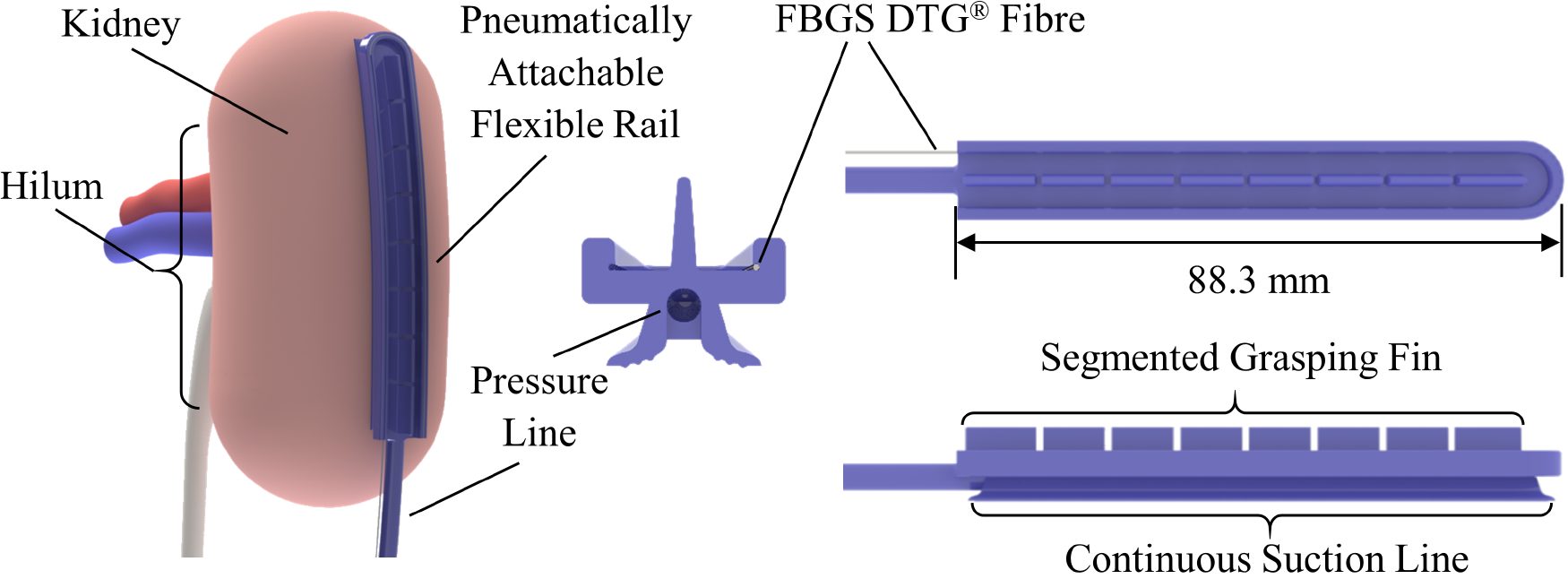}
    \caption{Schematic diagram of the PAF rail with integrated shape sensing fibre. The image on the left illustrates how the system could be attached to the kidney during RAPN. The images on the right illustrate the system design and the integration of the fibre sensing capability within the rail.}
    \label{fig:rail_kidney_schematic}
\end{figure*}

A potential aid for this sub-task is the deployment of soft robotic systems in the surgical workspace. Recently, the use of soft robots for minimally invasive surgery has gathered attention due to their inherent flexibility and compliance with their environment \cite{Runciman2019SoftSurgery}. In previous work \cite{Stilli2019PneumaticallyStudy}, we presented a Pneumatically Attachable Flexible rail (PAF rail) to enable stable, track-guided US scanning of the kidney during RAPN. The PAF rail is attached to the kidney surface using a continuous suction cup. Stable track-guided US scanning is achieved by connecting the drop-in US probe to the perimeter of the rail. In \cite{Ettorre2019Semi-AutonomousRails} the authors have investigated the autonomous deployment of the PAF rail on the surface of the organ and its use in intraoperative organ manipulation. In \cite{Wang2020UltrasoundStudy}, Wang et al. studied the 3D reconstruction of a mass embedded in a kidney phantom when the PAF rail guides the US probe. Accurate shape sensing of the PAF rail within the surgical field could further improve the deployment of this system on the targeted organ surface while autonomously controlling the probes' trajectory. 

The complete deployment process of the PAF rail and its use with US probes is detailed in \cite{Stilli2019PneumaticallyStudy}. However, deploying such a device during surgery presents many control challenges. Embedding sensors in the soft robot can give the information needed to meet said control challenges, but these sensors must have the ability to bend, twist, and contort in tandem with the soft robot.
In surgical soft robotics, the most common sensing methods are external vision-based systems using intra-operative imaging, \cite{ Wang2020UltrasoundStudy, Luo2015RobustFusion} and electromagnetic tracking, \cite{Luo2015RobustFusion}\cite{Lun2019Real-TimeGratings}, particularly in needle-based \cite{Hakime2012Electromagnetic-trackedResults} and catheter-based interventions \cite{Lugez2017ImprovedBrachytherapy}, \cite{Schwein2018ElectromagneticStudy}. But both these techniques present some drawbacks. Vision-based shape tracking is met with visual occlusion due to the constrained surgical workspace, while electromagnetic tracking is prone to extensive errors due to local field distortions.
Consequently, researchers have investigated the fibre-optics sensors in this context\cite{Sareh2015MacrobendArms}\cite{Silvestri2011Optical-FiberApplications}. These fibre-optic sensors are flexible, biocompatible, immune to electromagnetic interference and have small radial dimensions, making them ideal for surgical applications \cite{Mishra2011FiberApplications}. In particular, Fibre Bragg Grating (FBG)-based sensors allow direct multi-point strain measurements along the axial direction of the fibre and can contribute to real-time shape reconstruction \cite{Polygerinos2010MRI-compatibleProcedures}.

In this paper, we assess the performance of the real-time curvature and shape sensing of the PAF rail system using embedded FBG-based shape sensors. We further demonstrate how the PAF rail local shape can help plan a trajectory and autonomously guide an intraoperative US probe, thus having the potential to reduce surgeons' cognitive load while improving tumour margins identification. This can also improve the 3-D reconstruction accuracy of malignant masses while working towards being able to overlay intraoperatively the 3D reconstructed images in the field of view of the surgeon.

The paper is structured as follows; in section II the FBG-based curvature and shape sensing theory, and mechanical design and fabrication of the sensorised PAF rail is introduced. In section III, we show experimental studies evaluating the curvature sensing with the PAF rail applied to phantoms of different curvatures and materials. Different rail materials of various stiffness are also evaluated. We also demonstrate the ability to use the FBG-sensed shape to perform an autonomous US scan of a kidney phantom. Discussion of our findings is given in section IV, and finally, section V contains conclusions and future work.

\section{Materials \& Methods}
\subsection{FBG-based Shape Sensing}
FBG sensors are optical sensors that utilise Bragg reflection
to measure strain and temperature. FBG sensor fabrication requires altering an optical fibre with a laser and
an interferometer to draw a periodic change in the refractive index\cite{Bronnikov2019DurableFiber}, known as a grating. Each grating is wavelength-specific and, only a subset of the
light spectra, the Bragg wavelength, is reflected, while the rest is
transmitted. We can express this relationship as: 

\begin{equation} \lambda_B = 2 n_{eff}\Lambda\label{eq}\end{equation}
where $\lambda_B$ is the central wavelength of the reflected spectrum, $n_{eff}$ is the effective refractive index of the fibre core and $\Lambda$ is the grating pitch \cite{Zhuang2018FBGRobotics}.

The shape-sensor we use is a custom-made multi-core fibre (MCF) (CP-FBG DTG\textregistered{} (Draw Tower Grating), FBGS International, Jena, Germany), with a central core surrounded by seven equally spaced outer cores. Each core contains 25 FBGs spaced at 10 mm intervals along the optical axis, giving an overall sensing length of 240 mm. 
Bending of the shape-sensing fibre causes strains in the FBGs, in turn causing shifts in the Bragg wavelengths of the gratings, which are monitored by an optical interrogator. The raw wavelength data are converted to strains for each grating, and the strains of the four gratings at each position along the fibre are then used to compute to a local curvature at that position. The inital data acquisition and processing is performed by proprietary software (IllumiSense v3.1.x, FBGS  International,  Jena,  Germany) and a proprietary LabVIEW VI (National Instruments, Austin, TX, USA), and the curvature data is recorded and visualised through a custom Python application.

\subsection{Design and Fabrication of the Sensorized PAF rail}
The design of the PAF rail suction line used in this work is based on the optimisation study firstly presented in \cite{Stilli2019PneumaticallyStudy}. The structure of the PAF rail-tool interface is an improved version of the design proposed in \cite{Wang2020UltrasoundStudy}, where the customised grasping slot is replaced by a continuous segmented fin, enabling grasping of the PAF rail at any point along it. We added a  channel of 1 mm diameter along the internal perimeter of the rail to embed the shape sensing fibre, as shown in Fig. \ref{fig:rail_kidney_schematic}. 

The mould for this prototype was designed in SolidWorks (Dassault Systèmes, Vélizy-Villacoublay, France) and 3D printed with an Objet260 Connex (Stratasys, Eden Prairie, MN, USA) in VeroClear resin material. Liquid silicone was degassed in a vacuum chamber for ten minutes before being injected into the mould and left to cure at room temperature for the required period. 

We aimed to identify the silicone material that provided the most accurate shape sensing of the surface on which  PAF rail system is deployed while ensuring enough structural rigidity to mechanically pair with the modified US probe presented in \cite{Wang2020UltrasoundStudy}. To ensure robust pairing between the probe and the rail profile, a certain level of stiffness is required. We also hypothesized that PAF rails made of silicone significantly stiffer than the tissue targeted would locally deform it, while systems significantly softer would conform better but provide less accurate shape sensing. The goal of this part of the study was to identify the best trade-off between shape sensing accuracy, mechanical pairing and navigation of the US probe. 
To validate our hypotheses, we selected five silicones from the supplier Smooth-On Inc. (Macungie, PA, US), and we fabricated five PAF rails with different shore hardness: DragonSkin™ 10 NV (Shore hardness 10 A), DragonSkin™ 20 (Shore hardness 20 A), DragonSkin™ 30 (Shore hardness 30 A), Smooth-Sil™ 940 (Shore hardness 40 A) and Smooth-Sil 950™ (Shore hardness 50 A).

\begin{figure}
    \centering
    \includegraphics[width=0.49\textwidth]{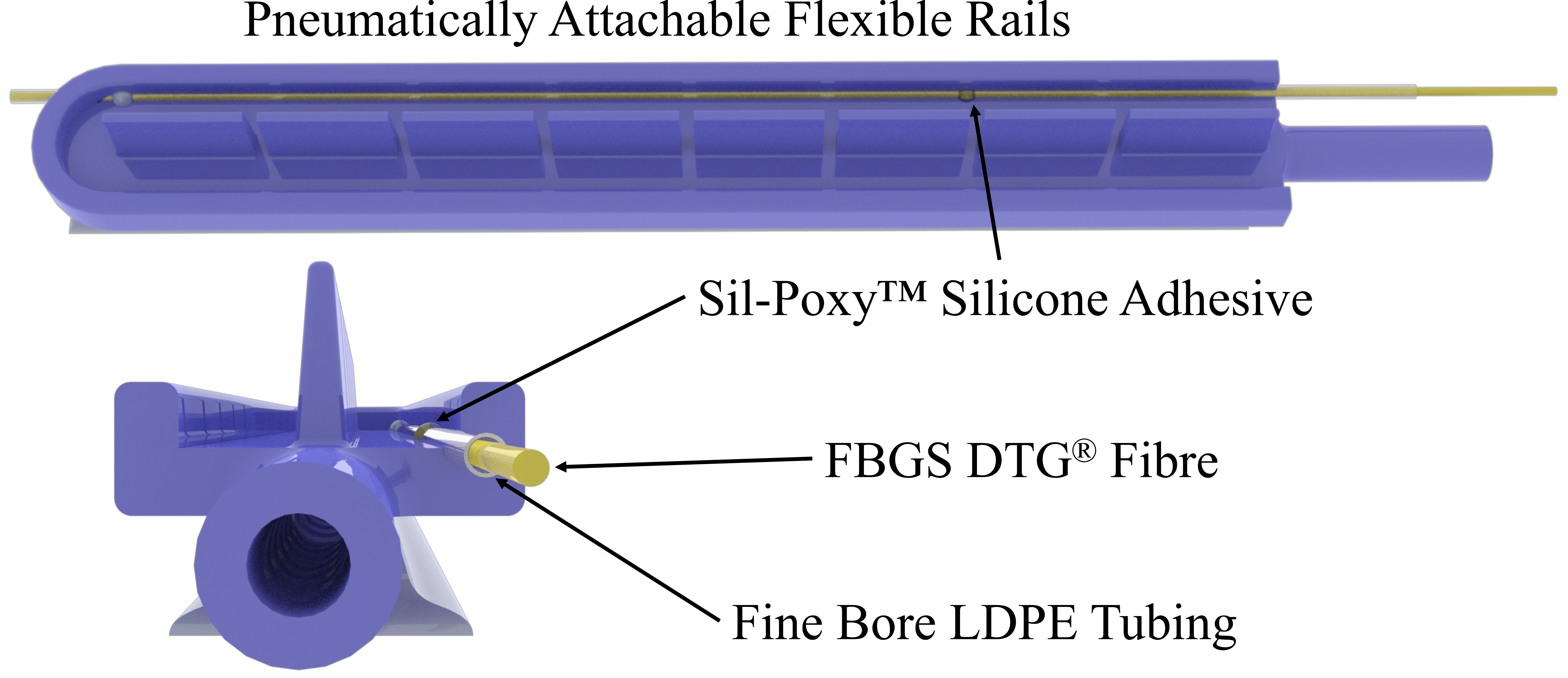}
    \caption{PAF rail with integrated Fine Bore LDPE tubing to house the FBGS DTG fibre. }
    \label{fig:PAF_rail_cross_section}
\end{figure}

The shape-sensing optical fibre is inserted in the rail through fine bore low-density polyethylene (LDPE) tubing (\O $0.86$ mm ID, \O $1.52$ mm OD). The tubing is fixed at the distal end of the PAF rail by pushing it through the silicone. Then, we secure it with Sil-Poxy™ silicone adhesive (Smooth-On Inc., Macungie, PA, US) at the proximal and distal ends, as depicted in Fig. \ref{fig:PAF_rail_cross_section}. 

Since the sensing length of the shape-sensing fibre is longer than the rail, only eight groups of four gratings lie within the rail profile, thus only these gratings are used to compute the PAF rail shape. The first grating group lies 3 mm from the rail's proximal end to guarantee that the curvature sensed by those gratings represents its current shape accurately. The fibre is secured to the tubing with Sil-Poxy™ silicone adhesive at the proximal end.

\subsection{Phantom Design and Fabrication}
To test the curvature sensing performance of the sensorized PAF rail, we fabricated several curvature phantoms of different curvature radii $R$ and materials. 

\begin{figure}[t]
    \centering
    \includegraphics[width=\columnwidth]{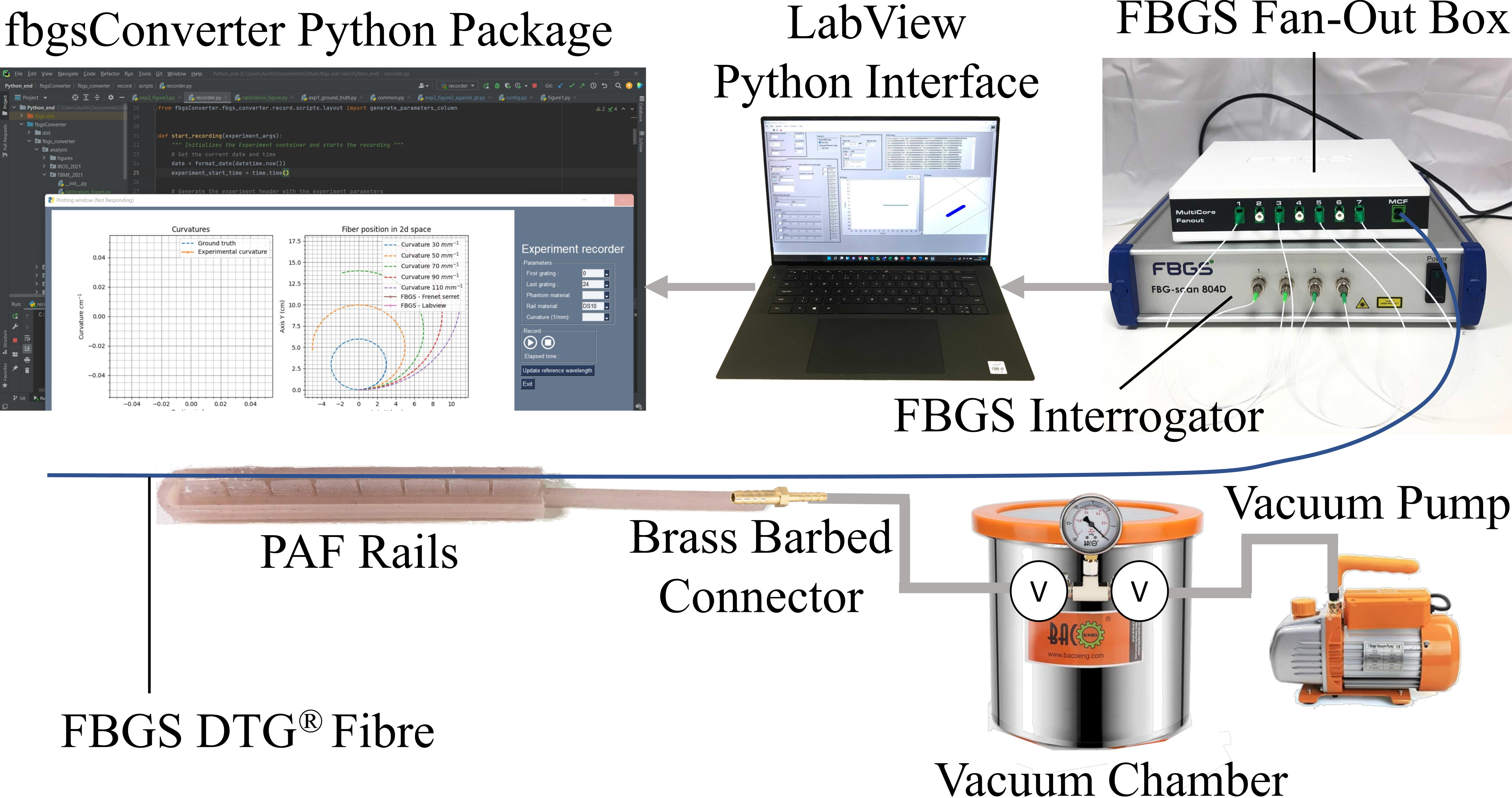}
    \caption{Schematic presenting an overview of the hardware and software elements of the proposed shape sensing system integration in the PAF rail system.}
    \label{fig:hardware_schematic}
\end{figure}

For the ground truth experiment, seven concentric circular grooves were laser cut into a sheet of acrylic plastic. The grooves each had a constant radius $R$, ranging from 30 mm to 110 mm in 20 mm increments. These measurements were chosen to correspond to the range of curvatures of an adult human kidney, as obtained from analysis of the KiTs19 dataset \cite{Heller2019TheOutcomes}\cite{Heller2021TheChallenge}. This will be referred to as the calibration plate for the remainder of this paper. The experimental setup is shown in Fig. \ref{fig:hardware_schematic}. 

\begin{figure}[b]
    \centering
    \includegraphics[width = \columnwidth]{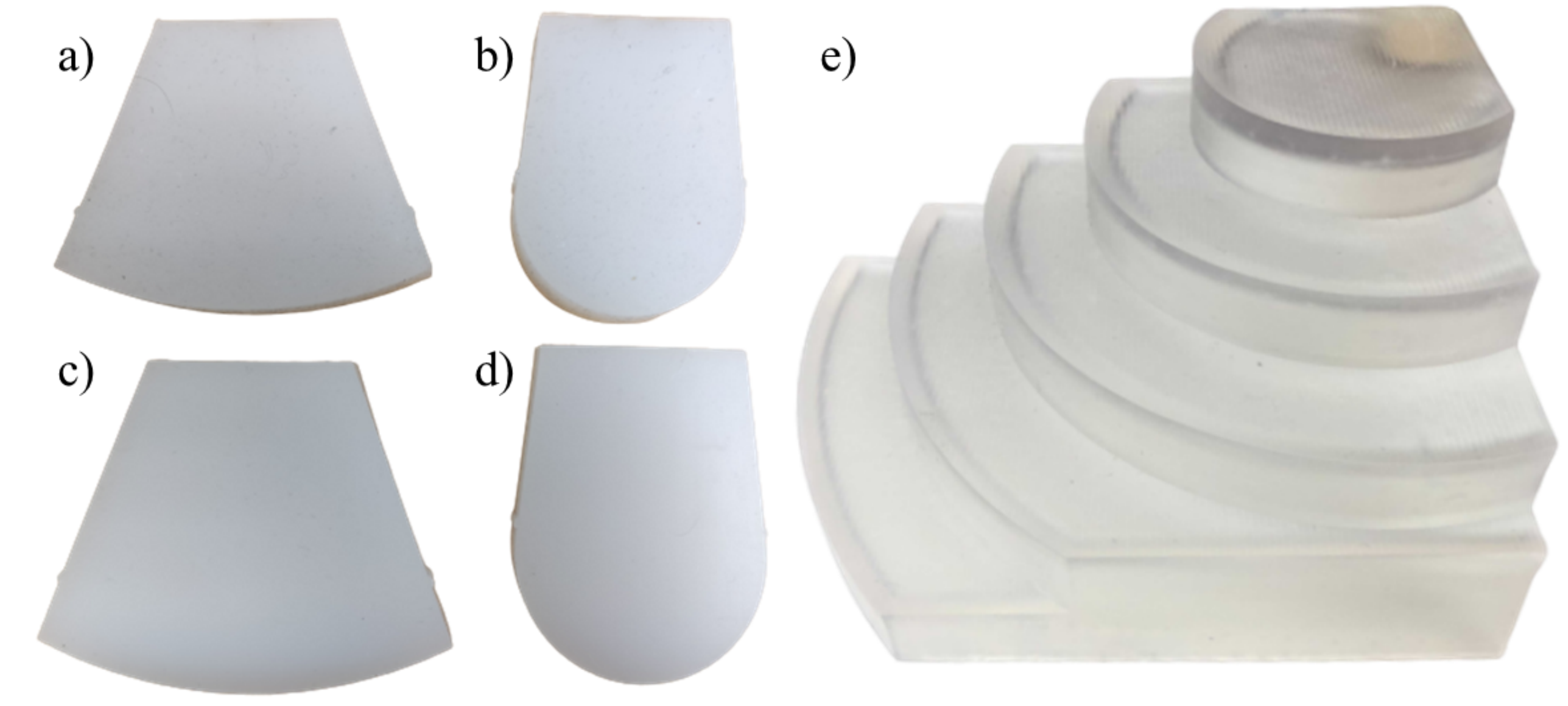}
    \caption{Soft curvature phantoms (material, curvature radius): a) EcoFlex\texttrademark{} 00-20, 110 mm b) EcoFlex\texttrademark{} 00-20, 30 mm c) DragonSkin\texttrademark{}, 110 mm d) DragonSkin\texttrademark{}30, 30 mm. Rigid curvature phantoms: e) 110 mm, 90 mm, 70 mm, 50 mm, 30 mm (ascending curvature). }
    \label{fig:phantoms}
\end{figure}

For experiments involving the PAF rail, we 3D printed a rigid curvature block comprised of seven different constant curvature surfaces. The curvatures range from 30 mm to 110 mm with 20 mm increments and,  each step has an elevation of 15 mm to accommodate for the width of the rail (Fig. \ref{fig:phantoms}(e)). 

\begin{table}[b]
\caption{Phantom Specifications}
\label{table:phantom_specifications}
\setlength{\tabcolsep}{3pt}
\begin{tabular}{p{70pt}p{70pt}p{95pt}}
\hline
Phantom Type& 
Material& 
Radius (mm) \\
\hline
Rigid & 
VeroClear Resin & 
30, 50, 70, 90, 110 \\
Soft & 
DragonSkin\texttrademark 3 & 30, 110 \\
Soft & 
EcoFlex\texttrademark 00-20& 
30, 110 \\
\hline
\end{tabular}
\label{tab1}
\end{table}

In addition, we fabricated four soft curvature phantoms with $R$ of 30 mm and 110 mm, each in DragonSkin™ 30 (Shore hardness 30A) and Ecoflex™ 00-20 (Shore hardness 00-20), Smooth-On Inc. (Macungie, PA, US) (Fig. \ref{fig:phantoms}(a-d)). These materials are suggested in \cite{Adams2017SoftSystem} and \cite{Cheung2014UsePyeloplasty} for the most realistic development of soft kidney phantoms. The phantom specifications are summarised in Table \ref{table:phantom_specifications}.

For the case study, we designed and fabricated an anatomically accurate kidney phantom with realistic mechanical properties, such as shape, volume and stiffness. The authors wish to acknowledge Dr. Efthymios Maneas and Prof. Adrien Desjardins for their help and guidance with the fabrication of this phantom. Using 3D Slicer \cite{Pieper20043DSlicer}, we reconstructed a 3D volume of an adult kidney using the CT images and semantic segmentation labels available from the KiTs19 dataset and made a negative mould in Clear Resin. Tissue-mimicking material (polyvinyl alcohol—PVA) was poured and cast within it. A spherical structure was fabricated and embedded within the phantom to replicate a malignant mass. We followed the same method used to make the kidney phantom but added a freeze-thaw cycle during the process \cite{Mackle2019Wall-lessVessels}. The additional freeze-thaw cycle increased the stiffness of the mass and the vasculature, thereby providing more realistic mechanical properties. 

\subsection{Ground Truth Curvature Sensing}
To assess the accuracy of the shape sensing fibre in combination with the optical system and IllumiSense software used for subsequent experiments, we evaluated the performance of the curvature sensing on a range of curvatures characteristic of a human kidney, as obtained from analysis of the KiTs19 dataset. We set the reference wavelengths by holding the fibre straight and flat against a parallel surface. Then, we positioned the sensing portion in each of the grooves of the calibration plate. The grooves held the fibre in place such that there was no need for external fixation. 
Wavelength shift and curvature were recorded at each grating index and for each curvature for 30 iterations. Then, the recorded curvatures were averaged over 30 data collected at a 100Hz rate for noise filtering. We repeated eight times the batch curvature measurements resetting the reference wavelength for each repetition of data collection. Since we experienced random noise during four recordings, we decided to label the defective recordings as outliers and removed them from the presented data.

\begin{figure}[t]
    \centering
    \includegraphics[width=\columnwidth]{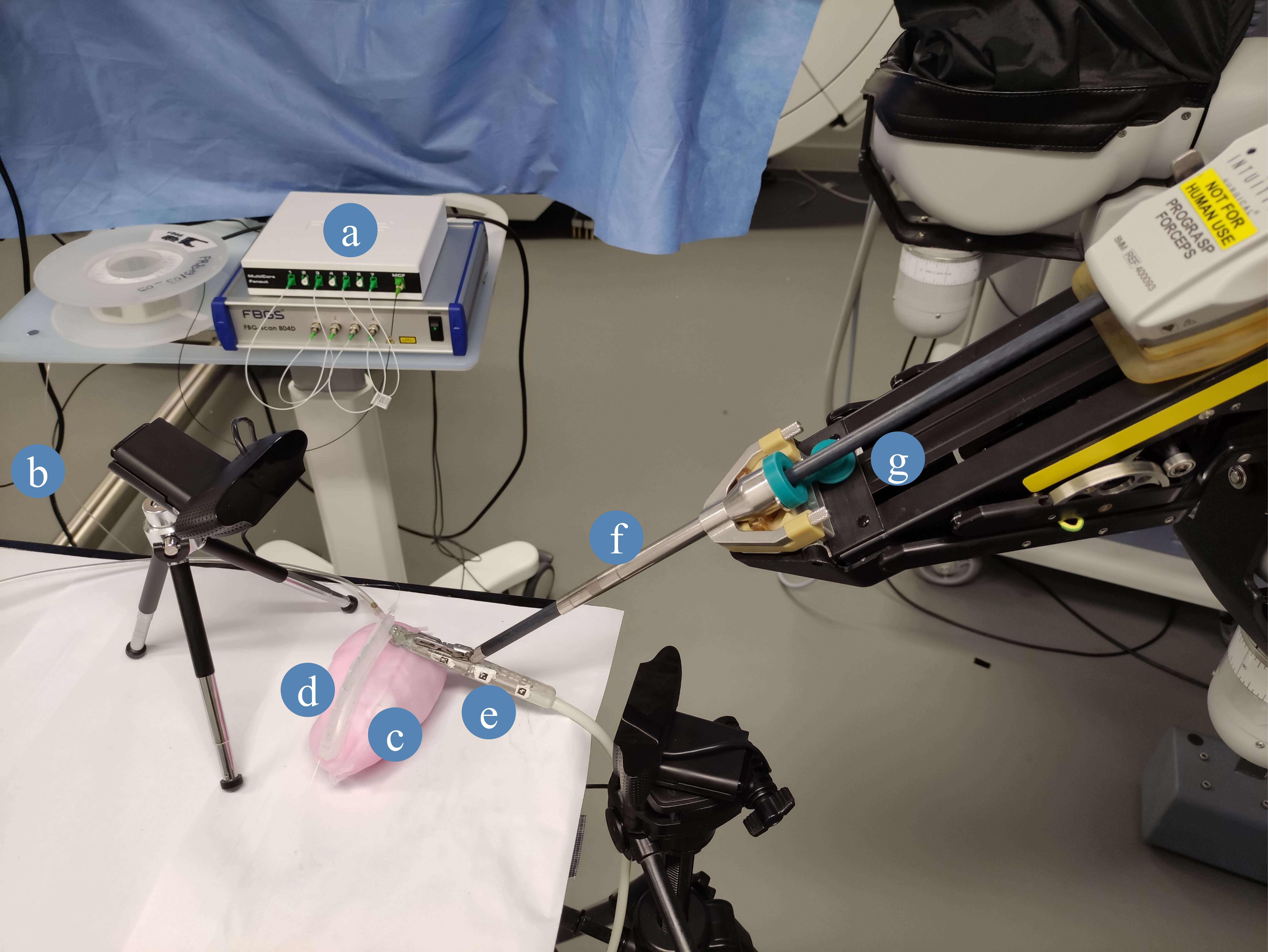}
    \caption{Case study experimental setup. a) FBGS Interrogator and Fan-Out Box: the fan-out is used to connect the multicore shape-sensing fibre to the optical interrogator. b) FBGS DTG fibre. c) PVA kidney phantom. d) DS30 PAF Rail. e) BK X12C4 Drop-In Ultrasound Probe. f) EndoWrist Prograsp Forceps, g) da Vinci surgical robot.}
    \label{fig:case_study_experimental_setup}
\end{figure}
 
\subsection{Curvature Sensing}
The fibre was embedded in each of the PAF rail samples as described in section II.B. We used a 3 CFM single-stage vacuum pump (Bacoeng, Hawthorne, CA) to vacuumize a 12-litres vacuum chamber (Bacoeng, Hawthorne, CA) and monitored the pressure thanks to an embedded manometer. The vacuum pressure used for all the tests was $P_{abs}=7.325 kPa$, as discussed in \cite{Stilli2019PneumaticallyStudy}. 
The chamber was connected with a pressure line to the tested PAF rail sample. Each of the five PAF rails was tested individually and, each test was repeated five times. 	

The PAF rail sample was suctioned to each of the curved surfaces of the rigid curvature block and soft curvature phantoms. The soft curvature phantoms were clamped in a vice to ensure that the stiffness of each material remained constant for different radii. Curvature data from the eight DTGs present within the rail was recorded for 30 iterations. The experiments were repeated five times.

\subsection{Stiffness Tests}
We conducted stiffness tests on porcine kidney tissue samples, silicone samples from the supplier Smooth-On Inc. and a PVA sample to select the material closest to the stiffness of an adult human kidney and to quantify the relative stiffness differences between the rail and the target organ. 
Using a surgical scalpel and a template, we excised cylindrical kidney tissue samples from the thickest part of each kidney. We fabricated silicone samples of the same geometry using a mould 3D printed in Clear Resin. The PVA sample was cut out of a PVA block to mitigate shrinkage when moulding.
The geometrical properties of each kidney and silicone sample are summarised in Table \ref{table:experimental_material_stiffness}. We conducted the stiffness tests using a UR3e robot arm (Universal Robots, Odense, Denmark), to which we attached a Mini 40 Force Sensor (ATI Industrial Automation, NC, USA) with a custom mount. The mount also included a custom indenter. Then, we planned trajectories and recorded position and force data with ROS.

\begin{figure}[b]
    \centering
    \includegraphics[width=\columnwidth]{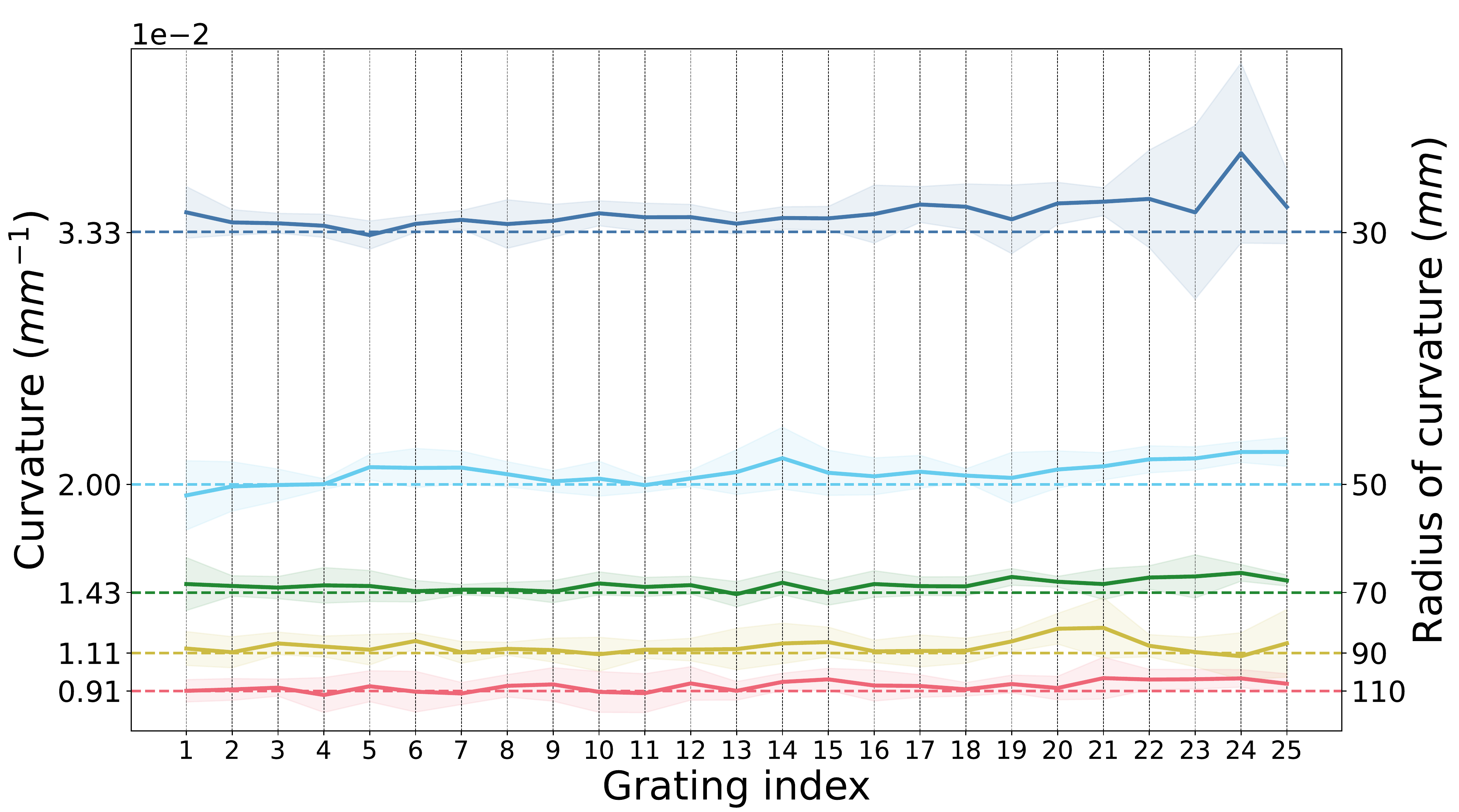}
    \caption{Ground Truth Sensed Curvature at each grating. Mean sensed curvature (solid line) against groove curvature (dashed line) at each grating index along the fibre. The shaded region represents the standard deviation over five iterations of recordings.}
    \label{fig:bare_fiber_curvature_sensing}
\end{figure}

\subsection{Case Study}
In a robotic-assisted partial nephrectomy, the surgeon must identify the tumour margins before the excision. To do so, the surgeon scans the kidney with a drop-in intraoperative US probe paired with either the EndoWrist\textregistered{} Prograsp\texttrademark{} or the Large Needle Driver (LDN) of the da Vinci Surgical System (Intuitive Surgical Inc., Sunnyvale, CA, US). It is a challenging task that can only be achieved by a highly-skilled surgeon. 

As a proof-of-concept of the fibre integration in the PAF rail for path planning, we automated this US scan. The robotic system employed to conduct this demonstration is the first generation da Vinci robot alongside the da Vinci Research Kit (dVRK) platform \cite{Kazanzidesf2014AnSystem}\cite{DEttorre2021AcceleratingKit}. We equipped the first da Vinci Patient Side Manipulator (PSM1) with the EndoWrist\textregistered{}  Prograsp\texttrademark{} Forceps paired with the drop-in US probe BK X12C4 (BK-Medical Holding Inc., Peabody, Massachusetts), as shown in Fig. \ref{fig:case_study_experimental_setup}.

Among the PAF rail systems manufactured in different materials, we selected the one made of DragonSkin\texttrademark 30 for this specific testing because it provides the best trade-off in terms of adhesion performance on hard surfaces and ability to withstand high vacuum pressure without collapsing. Further findings are discussed in section III.B.
The case study we are presenting in this section can be summarised with the following steps (See also Fig. \ref{fig:case_study_experimental_setup}):

\begin{enumerate}
\item The fibre is embedded in the DragonSkin\texttrademark 30 rail. 
\item Suction is achieved by applying vacuum pressure on the line (as described in III.C) to attach the PAF rail to the kidney phantom.
\item The drop-in US probe is manually paired with the EndoWrist\textregistered{} Prograsp\texttrademark{} forceps gripper of the da Vinci Surgical System itself mounted on one of the dVRK arms.
\item The PSM1 is manually positioned so that the connector at the tip of the drop-in US probe is paired with the rail profile perpendicularly to the fibre in correspondence of the first grating.
\item We take an instantaneous reading of the 2-D shape of the optical fibre generated by the FBGS proprietary software. This shape is computed out of the spectra of the fibre recorded through the hardware connection with the FBGS Fan Out Box and Interrogator. 
\item The 2-D shape is converted into a 3-D point trajectory, assuming no displacement along the z-direction since we paired the probe perpendicular to the PAF rails. And publish to the dVRK computer through ROS.

\item Then, the 3-D trajectory is converted to a PosedStamped message in  ROS. Here, we control the frame called "PSM1\_psm\_base\_link", which corresponds to the tooltip coordinates. First, we extract the current frame. For each position along the trajectory, we compute the novel frame by adding the 3-D coordinates to the current frame. 

\item Finally, the poses are published to the dVRK. The US probe moves along the PAF-rail following said trajectory and generate US images of the phantom.
\end{enumerate}

\section{Results $\&$ Discussion}
\subsection{Ground Truth Curvature Sensing}
Fig. \ref{fig:bare_fiber_curvature_sensing} shows the sensed curvature against geometric curvature for each of the curvature grooves in the calibration plate at each grating. Sensing accuracy increases as the curvature is reduced while variation along the fibre (for each grating) also decreases, giving an overall better curvature measurement at larger radii.

The absolute and $\%$ errors to geometric curvature are summarised in Table \ref{table:full_bare_fibre_error} for the entire length of the fibre (gratings 1 to 25 inclusive) and in Table \ref{table:subset_bare_fibre_error} for a subset of the gratings (gratings 11 to 19 inclusive).

\begin{table}[h]
\centering
\caption{Bare Fibre Sensed Curvature Errors (Gratings 1 to 25)}
\label{table:full_bare_fibre_error}
\setlength{\tabcolsep}{3pt}
\begin{tabular}{ccccc}
\hline
$R$  (mm) & 
Max $\epsilon$ ($\times 10^{-2} mm^{-1}$) & 
 ($\%$) & 
 Mean $\epsilon$ ($\times 10^{-2} mm^{-1}$) 
 &  ($\%$) \\
\hline
30&
$0.42 \pm 0.48$ & 12.5 &
$0.1 \pm 0.11$ & 2.9\\

50&
$0.17 \pm 0.08 $ & 8.6 &
$0.06 \pm 0.04$ & 3.2\\

70&
$0.1 \pm 0.04 $ & 7.3 &
$0.04 \pm 0.03$ & 2.9\\

90&
$0.13 \pm 0.16 $ & 11.9 &
$0.03 \pm 0.03$ & 2.9\\

110&
$0.07 \pm 0.11 $ & 7.6 &
$0.03 \pm 0.02$ & 2.8\\

\hline
\end{tabular}
\end{table}

\begin{table}[h]
\centering
\caption{Bare Fibre Sensed Curvature Errors (Gratings 11 to 19)}
\label{table:subset_bare_fibre_error}
\setlength{\tabcolsep}{3pt}
\begin{tabular}{ccccc}
\hline
$R$  (mm) & 
Max $\epsilon$ ($\times 10^{-2} mm^{-1}$) & 
 ($\%$) & 
 Mean $\epsilon$ ($\times 10^{-2} mm^{-1}$) 
 &  ($\%$) \\
\hline
30&
$0.39 \pm 0.78$ & 1.00 
 &  $0.73 \pm 0.37$ & 2.43\\

50&
$0.11 \pm 0.70 $ & 0.22 &
$1.36 \pm 1.02$ & 2.72\\

70&
$1.21 \pm 2.47 $ & 1.73 &
$1.21 \pm 0.82$ & 1.74\\

90&
$3.73 \pm 8.89 $ & 4.14 &
$2.27 \pm 1.67$ & 2.52\\

110&
$7.13 \pm 13.9 $ & 6.48 &
$1.59 \pm 4.18$ & 1.44\\

\hline
\end{tabular}
\end{table}

\begin{figure}[b]
    \centering
    \includegraphics[width=\columnwidth]{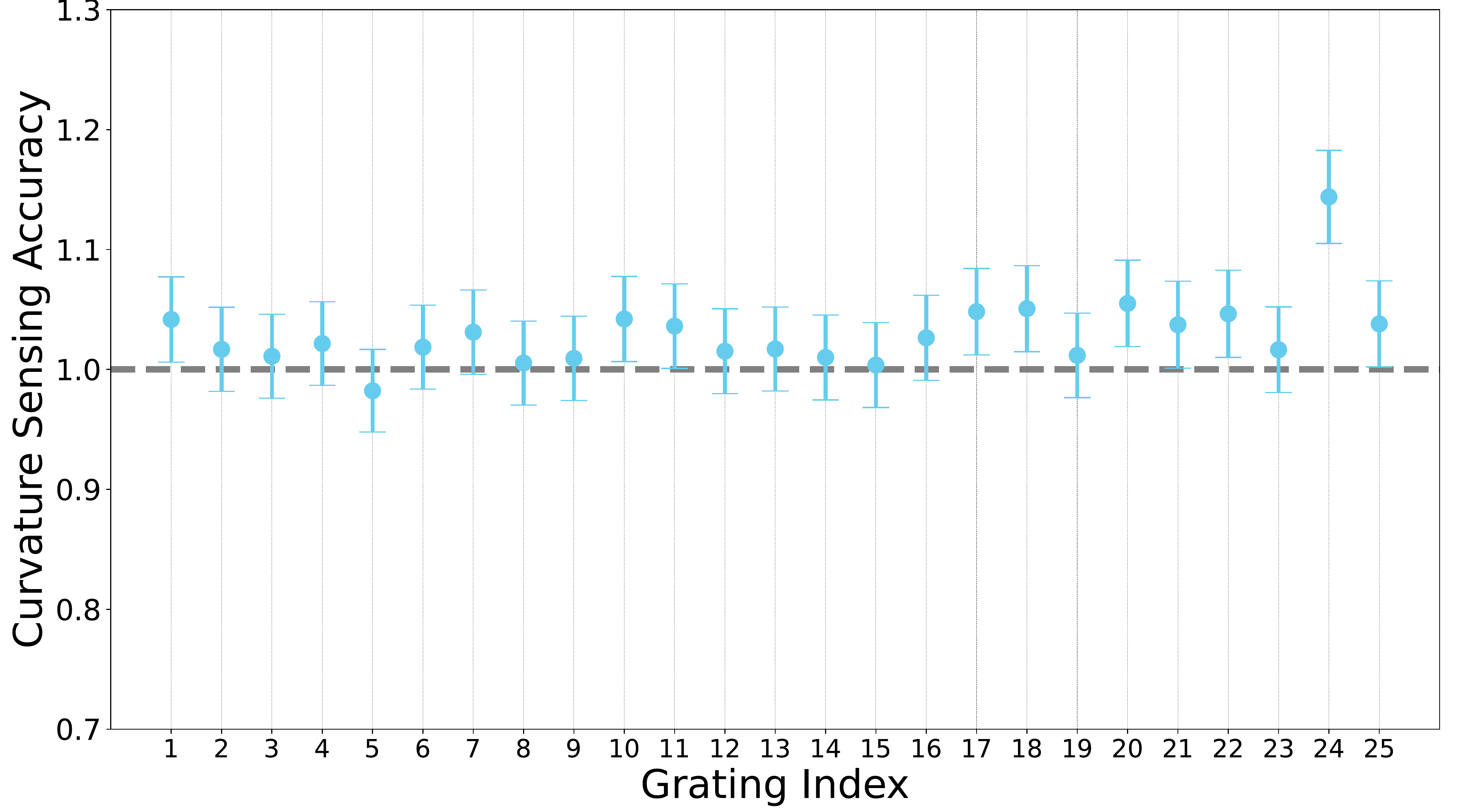}
    \caption{Curvature sensing accuracy of each grating along the fibre. Accuracy is defined as the ratio of sensed curvature to geometric curvature, for perfect sensing we would expect a value of 1 a each grating.}
    \label{fig:gratings_curvature_sensing_accuracy}
\end{figure}

We further evaluated the curvature sensing accuracy of each grating by calculating the ratio of sensed curvature to geometric curvature, averaged over the curvature range.  We expect a value of 1 for perfect sensing accuracy (shown in Fig. \ref{fig:gratings_curvature_sensing_accuracy}). 

Overall, the average sensing error (average of all relative errors at all grating indices at all radii)  is $1.02 \pm 0.03$ $(2.9\%)$. This is comparable to curvature accuracy reported in \cite{Zhuang2018FBGRobotics}. 

\begin{figure}[!tbp]
  \centering
  \begin{minipage}[b]{0.4\textwidth}
    \includegraphics[width=\textwidth]{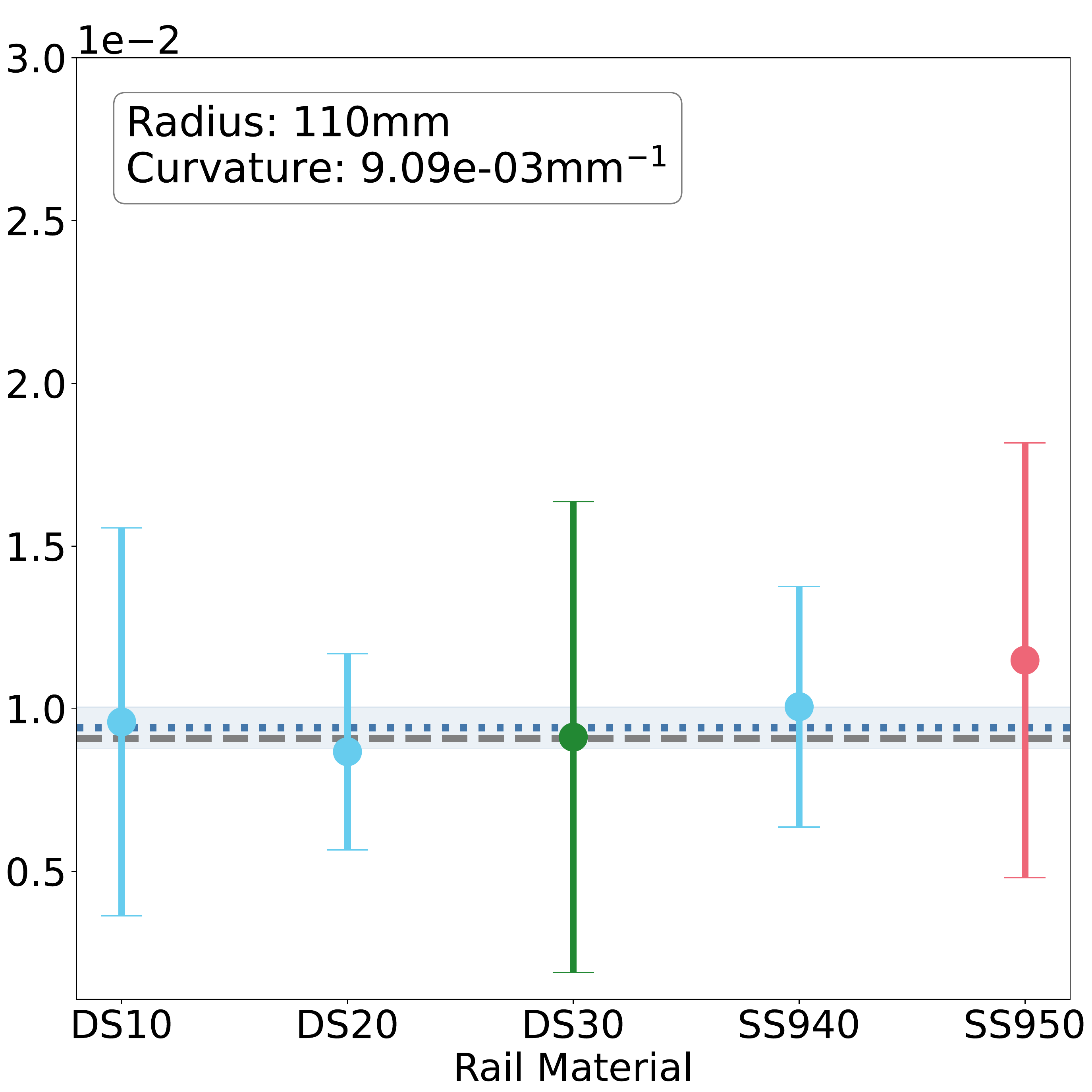}
    % \caption{Flower one.}
  \end{minipage}
  \hfill
  \begin{minipage}[b]{0.4\textwidth}
    \includegraphics[width=\textwidth]{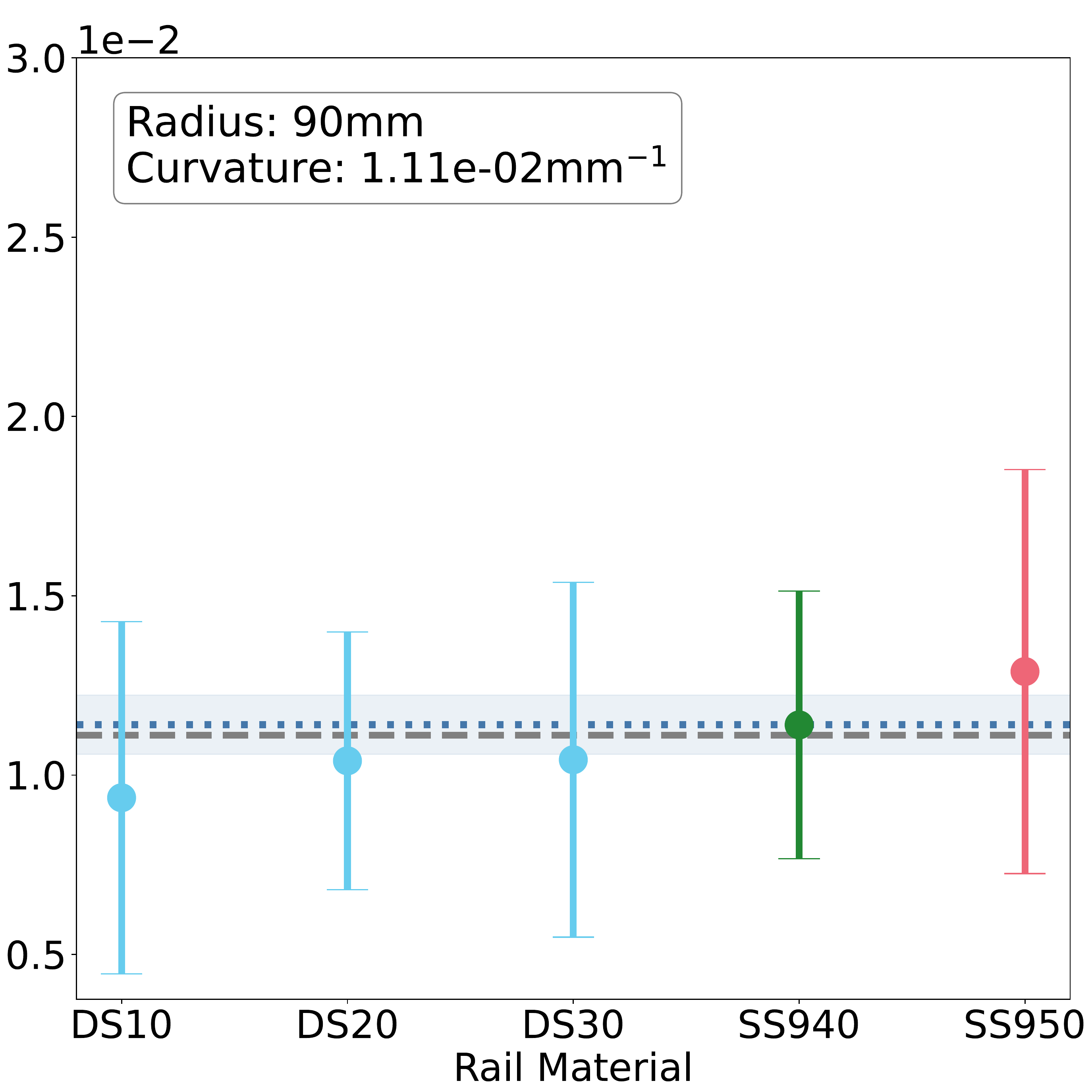}
    % \caption{Flower two.}
  \end{minipage}
  \hfill
  \begin{minipage}[b]{0.4\textwidth}
    \includegraphics[width=\textwidth]{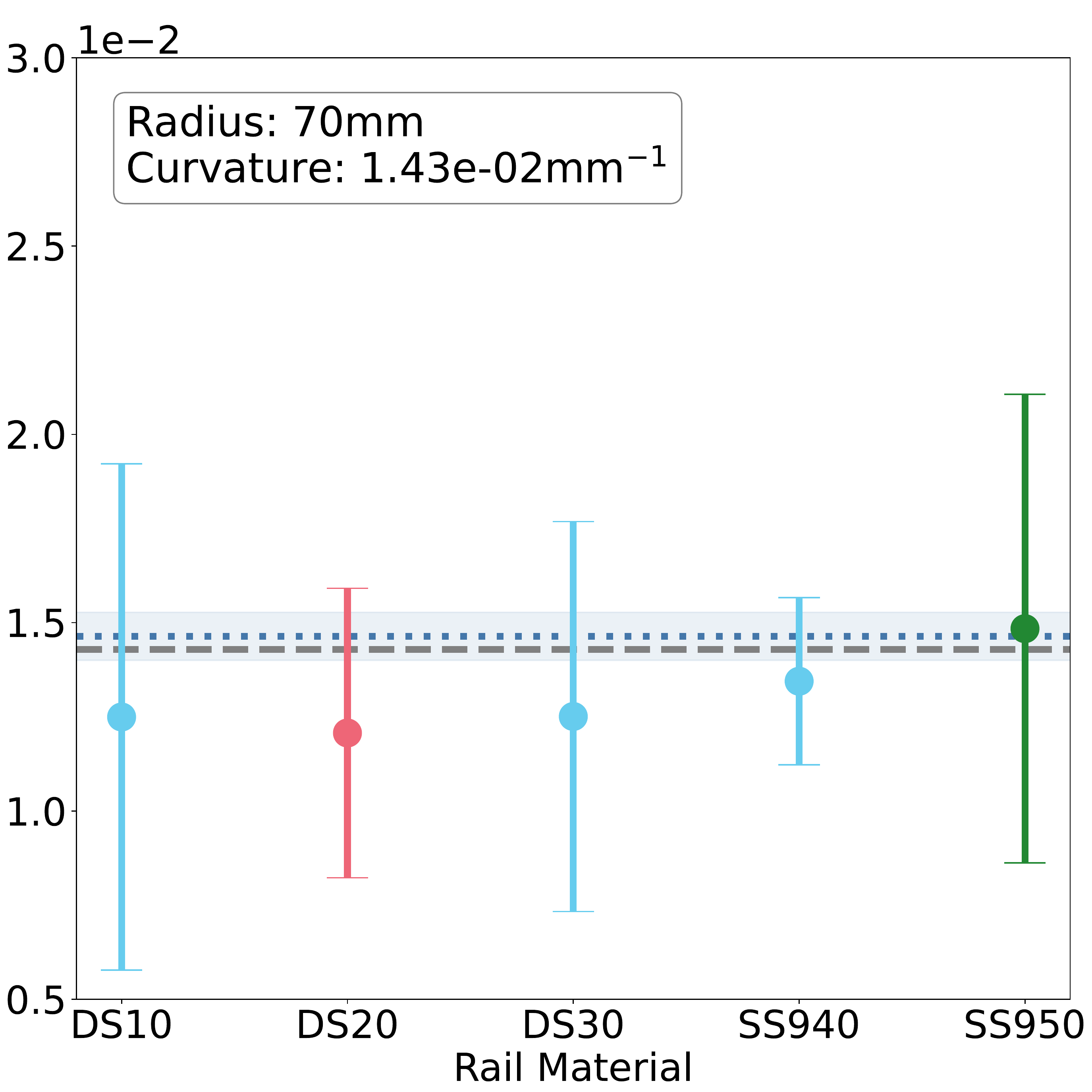}
    % \caption{Flower two.}
  \end{minipage}
  \hfill
  \begin{minipage}[b]{0.4\textwidth}
    \includegraphics[width=\textwidth]{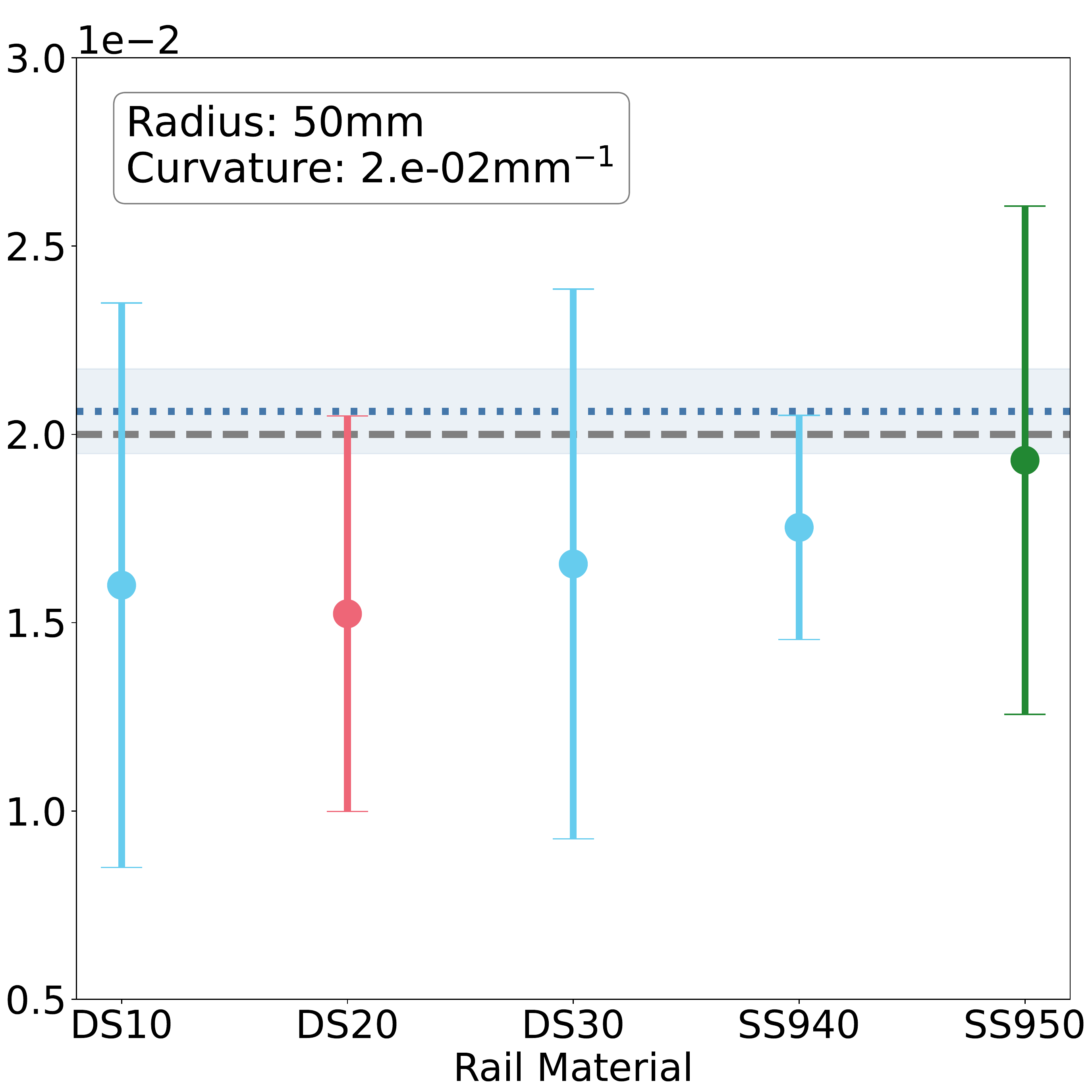}
    % \caption{Flower two.}
  \end{minipage}
  \hfill
  \begin{minipage}[b]{0.4\textwidth}
    \includegraphics[width=\textwidth]{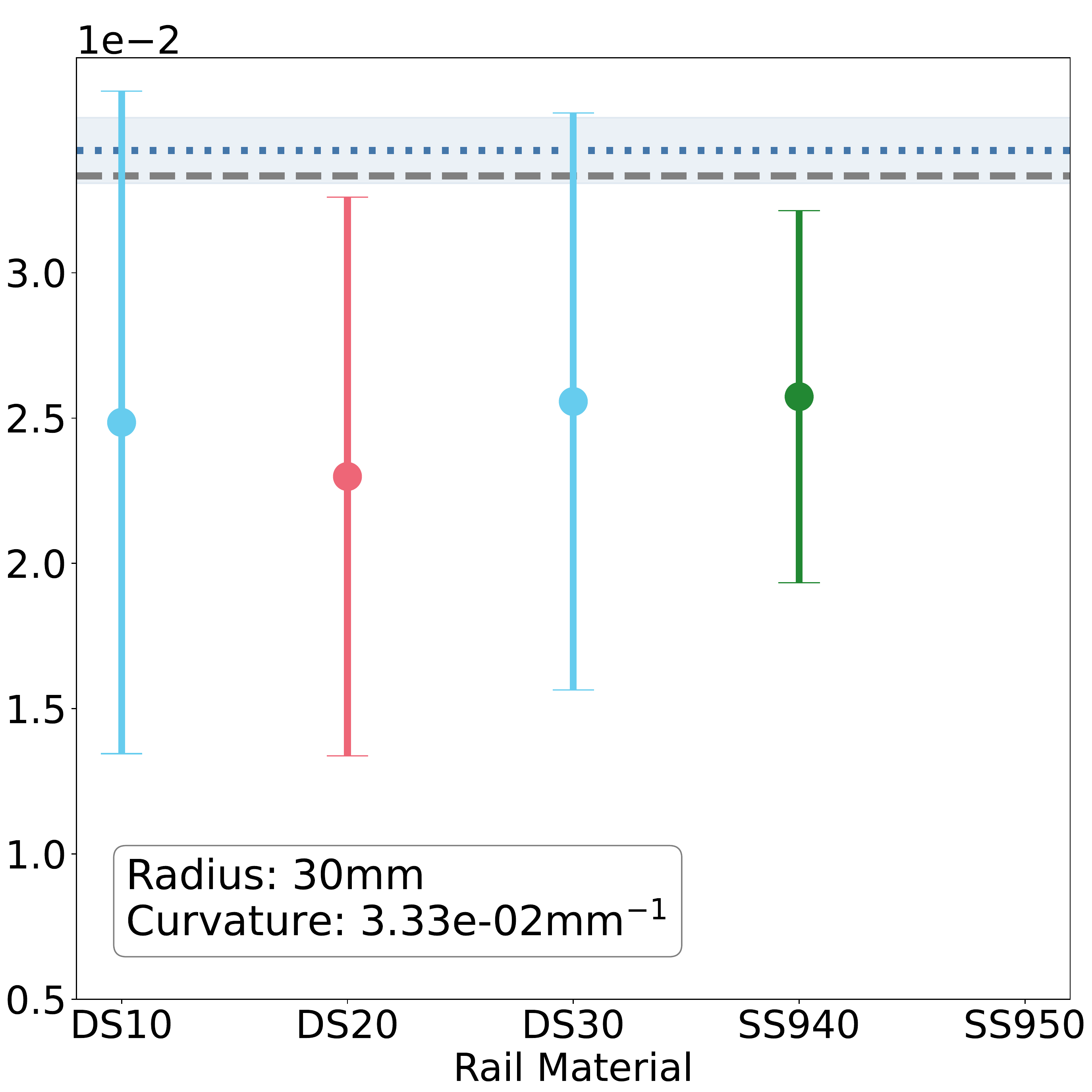}
    % \caption{Flower two.}
    \end{minipage}
    \hfill
  \begin{minipage}[b]{0.4\textwidth}
    \includegraphics[width=\textwidth]{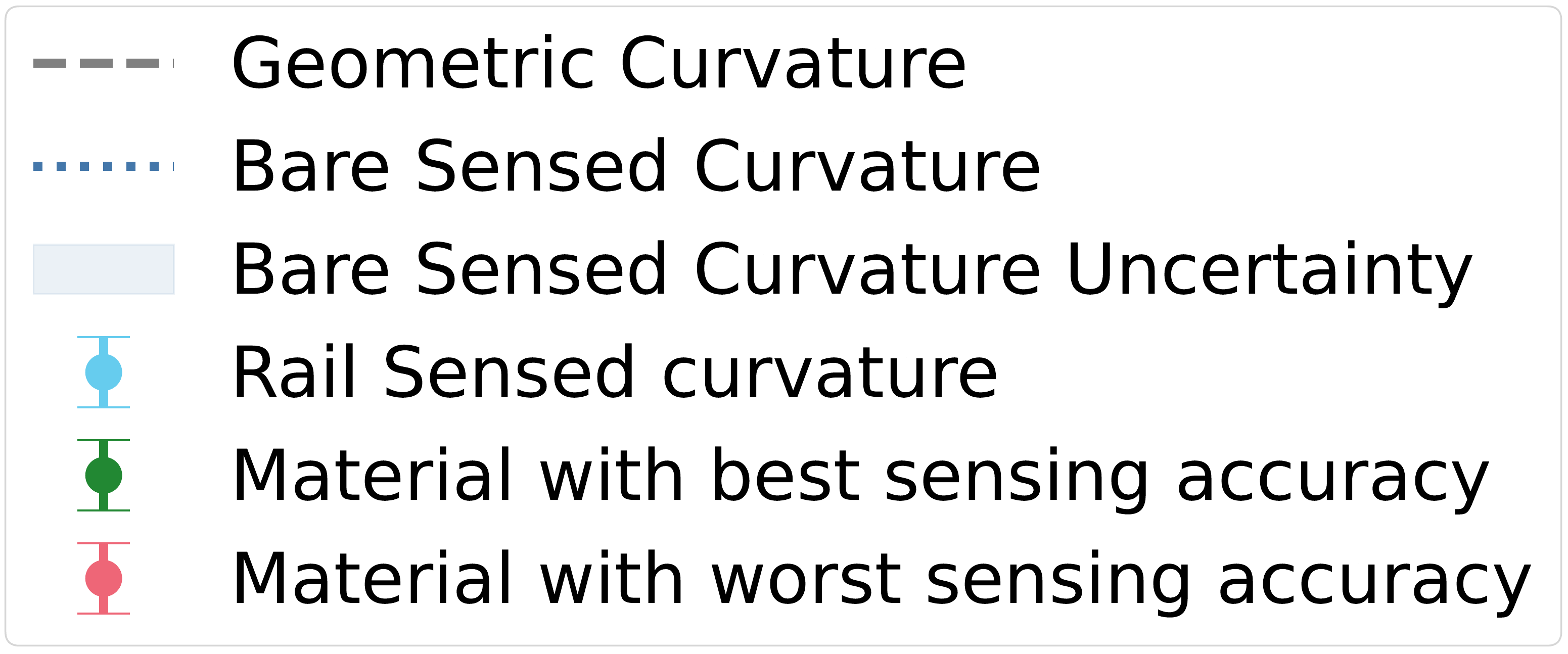}
    % \caption{Flower two.}
  \end{minipage}
  \caption{Rail-sensed curvature at each radius (decreasing L-R) on the rigid phantom. The materials on the $x$-axis are in order of increasing material stiffness. Green corresponds to the rail with best sensing accuracy, red to the worst.}
  \label{fig:rigid_phantom_results}
\end{figure}

\begin{figure}[!tbp]
  \centering
  \begin{minipage}[b]{0.4\textwidth}
    \includegraphics[width=\textwidth]{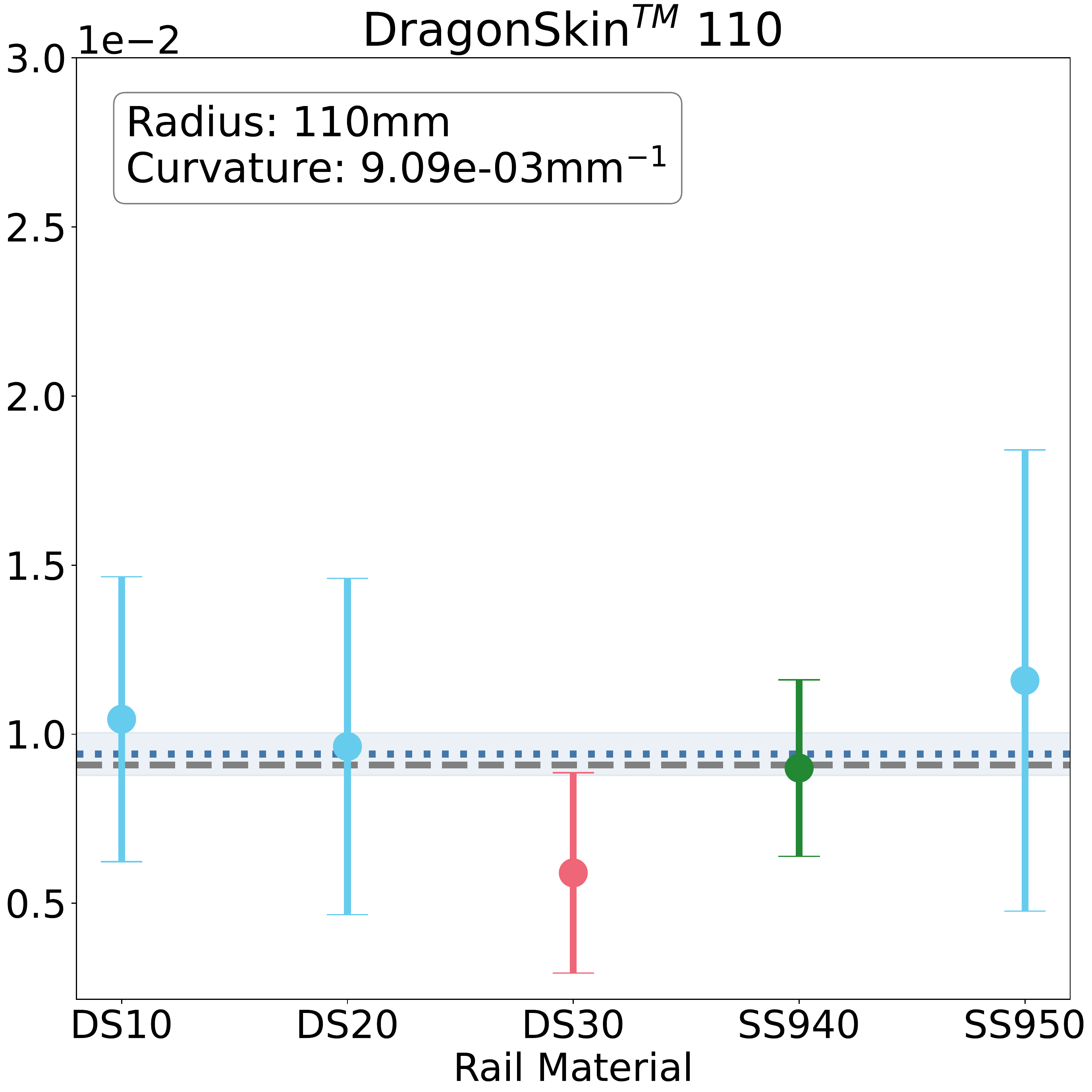}
    % \caption{Flower one.}
  \end{minipage}
  \hfill
  \begin{minipage}[b]{0.4\textwidth}
    \includegraphics[width=\textwidth]{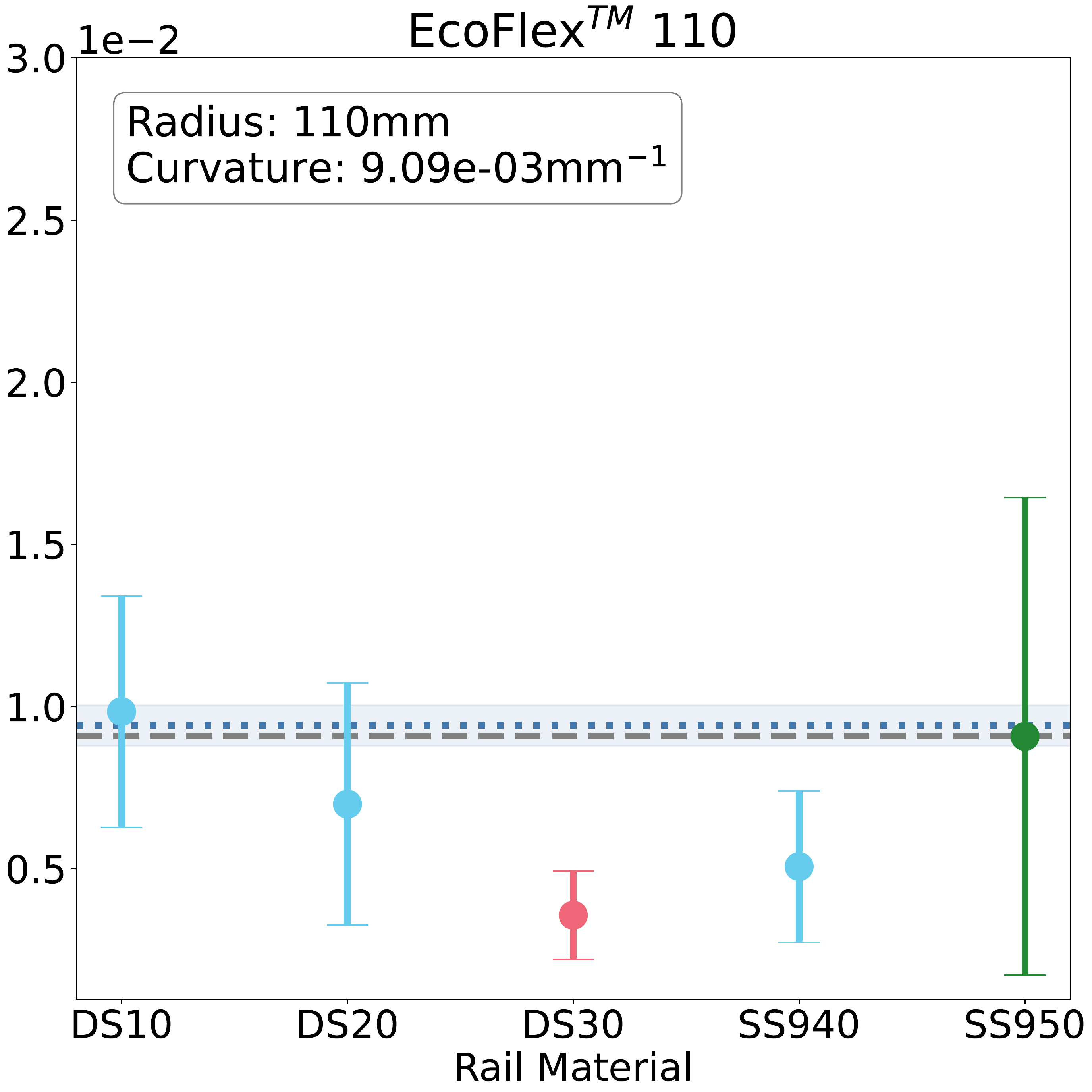}
    % \caption{Flower two.}
  \end{minipage}
  \hfill
  \begin{minipage}[b]{0.4\textwidth}
    \includegraphics[width=\textwidth]{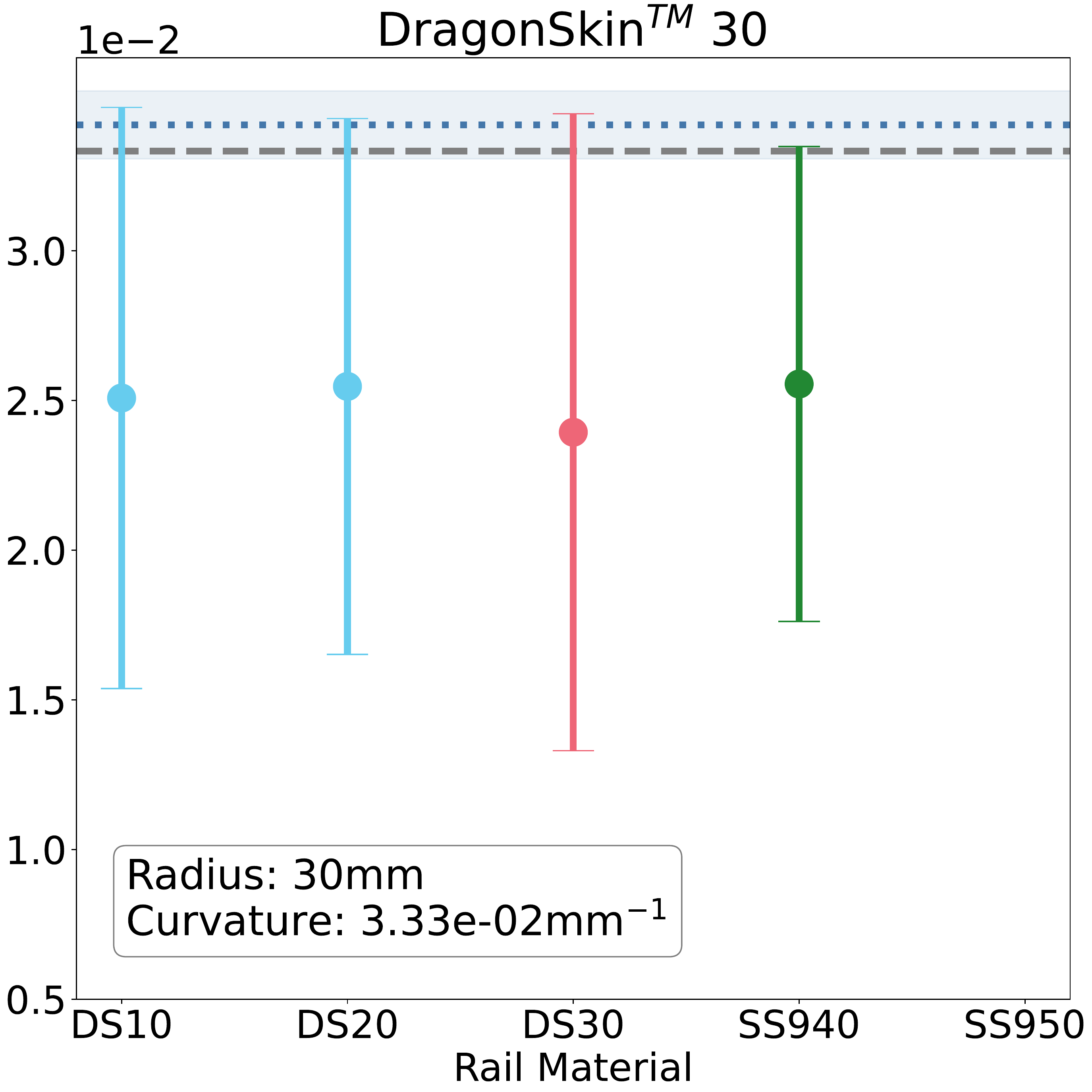}
    % \caption{Flower two.}
  \end{minipage}
  \hfill
  \begin{minipage}[b]{0.4\textwidth}
    \includegraphics[width=\textwidth]{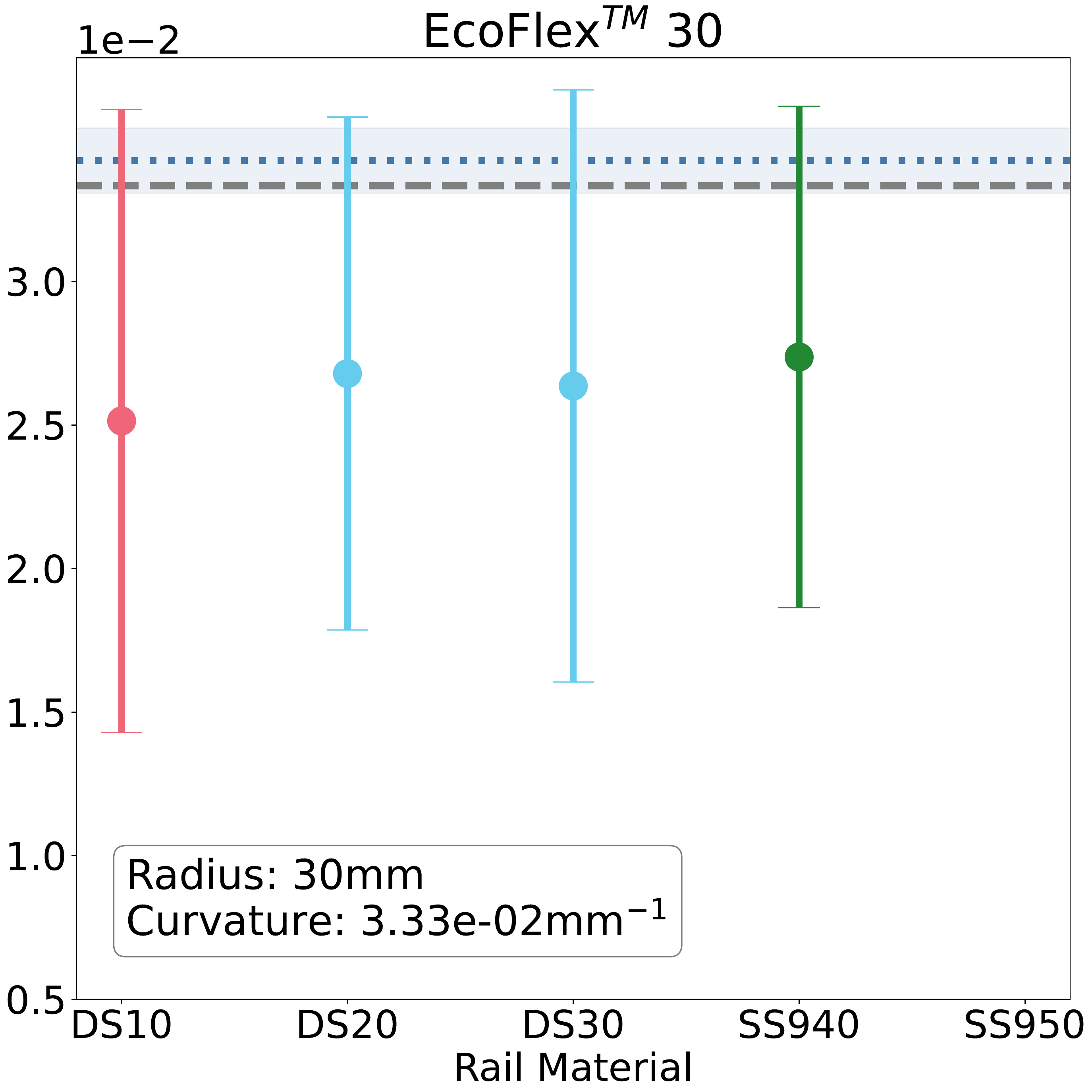}
    % \caption{Flower two.}
  \end{minipage}
  \caption{Rail-sensed curvature at $110$ mm (top) and $30$ mm (bottom) on the soft phantoms: DragonSkin\texttrademark (L) and EcoFLex\texttrademark (R). The materials on the $x$-axis are in order of increasing material stiffness. Green corresponds to the rail with best sensing accuracy, red to the worst.}
  \label{fig:soft_phantom_results}
\end{figure}
\subsection{Curvature Sensing}
Fig. \ref{fig:rigid_phantom_results} shows the average sensed curvature of each rail material for each radius compared with the ground-truth sensed curvature and geometric curvature when tested with the rigid phantom. The rail sensed curvature accuracy deteriorates as radius is decreased, agreeing with the bare fibre sensing behaviour. At $R = 110$ mm the best rail is DS30 while the worst is SS950. At $R = 90$ mm the best rail is SS940 while the worst is SS950. At $R = 70$ mm the best rail is SS950 while the worst is DS20. At $R = 50 $ mm and $R = 30 $ mm overall sensing accuracy is significantly worse, however SS950 and SS940 are the best performers respectively, while DS20 is the worst at both radii. Furthermore, each datapoint has a large uncertainty arising from the large variances in the raw data. 

\begin{figure}[b]
    \centering
    \includegraphics[width=\columnwidth]{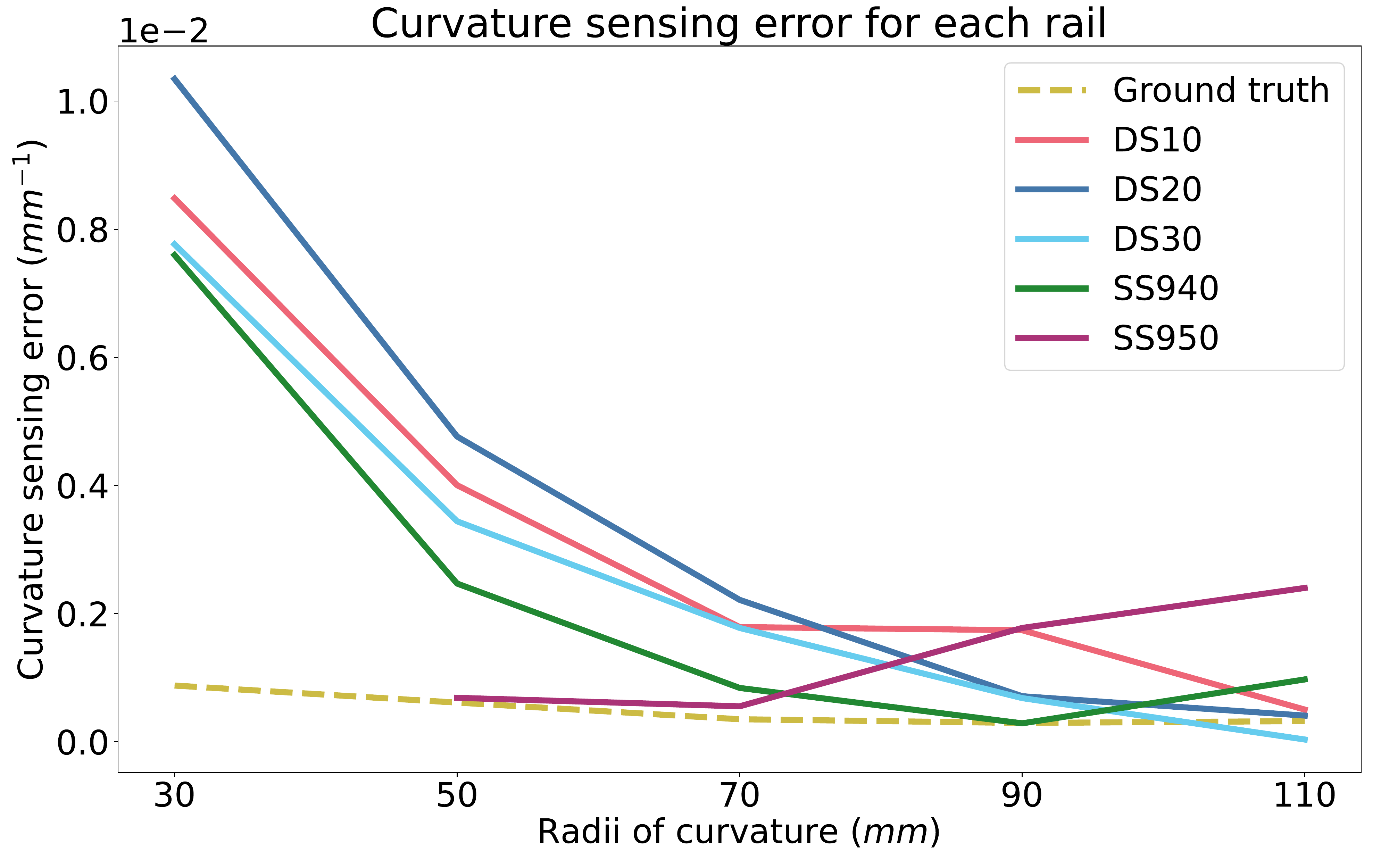}
    \caption{Absolute curvature sensing error for each rail material over the radii tested on the rigid phantom. Ground truth curvature sensing error is added as a baseline.}
    \label{fig:rigid_rail_analysis}
\end{figure}

In this experiments, we studied the influence of the rail material over curvature sensing performance.  The idea behind these experiments was to select a suitable material for the task that provided the best curvature sensing performance. 

Thus, to improve the readability of the results presented in Fig. \ref{fig:rigid_phantom_results} and Fig. \ref{fig:soft_phantom_results}, we coloured in green the rail expressing the lowest absolute error between measured and ground truth curvature. On the contrary, we highlighted the worst-performing rail in red.      
  
Fig. \ref{fig:rigid_rail_analysis} shows that there is a trend linking the curvature sensing accuracy to tested curvature. At smaller curvatures, the rails prove to be more accurate than at larger curvatures. It contradicts the results obtained during the ground truth curvature sensing presented in Fig. \ref{fig:bare_fiber_curvature_sensing}. One hypothesis to explain this trend is the effect of the stiffness of the optical fibre exceeding that of the rail. This mismatch in stiffness properties may cause the fibre to bend tangentially to the rail at smaller radii (50 mm and 30 mm), as opposed to concentric to it. Further experiments could ascertain the stiffness of the fibre to compare with the rail, however this is out of the scope of this study. 

In Fig. \ref{fig:rigid_phantom_results}, DS20 and SS950 present the worst sensing accuracy for two or more curvatures and SS940 had the best curvature sensing performance on the rigid phantom. DS30 had the second best performance.

Subsequently, we evaluated the curvature sensing performance on soft phantoms that have similar material properties to that of kidney tissue. 
Results for the soft curvature phantoms for each rail material are shown in Fig. \ref{fig:soft_phantom_results}. For this experiment, only the extremities of the range of radii were tested. Again, all figures show the error to ground-truth sensed curvature and geometric curvature. Fig. \ref{fig:soft_phantom_results} shows the results for phantoms made in DragonSkin\texttrademark{} 30 silicone (Left), and Fig. \ref{fig:soft_phantom_results} (R) shows that for EcoFlex\texttrademark{} 00-20 silicone.

In agreement with both the bare fibre data and the rigid sensing data, we see that sensing accuracy is better $R = 110$ mm than at $R = 30$ mm. At $R = 110$ mm DS30 is the worst rail material for both DragonSkin\texttrademark{} and EcoFlex\texttrademark{} phantoms. In the former phantom material, SS940 is the best rail material while in the latter phantom material, SS950 is the best rail material. 
At $R = 30$ mm all rail materials exhibit similar poor performance ($> 25$\% error to geometric curvature). There is no data for the SS950 rail on the EcoFlex\texttrademark{} phantom as the rail could not stick to this radius for the required period. 
The softer rails tend to collapse under vacuum pressure. As a result, they tend to express which can explain lower curvature sensing accuracy. Moreover, stiffer rails were harder to stick to the phantom surface. Especially with SS950, we could not maintain the contact at $R = 30$ with the phantoms for the experiment duration.

Current experiments suggests that SS940 proved better curvature sensing over all the tested curvatures and phantoms as depicted in Fig. \ref{fig:rigid_rail_analysis}. However, it was hard to stick it to the surface of the rigid curvature block and this material was not selected for the experiments presented in the next section. Thus, since the soft phantoms did not highlighted any significant trend and that DS30 was the second best material on the rigid curvature, we selected it for the next experiment.

\begin{figure}
    \centering
    \includegraphics[width=\columnwidth]{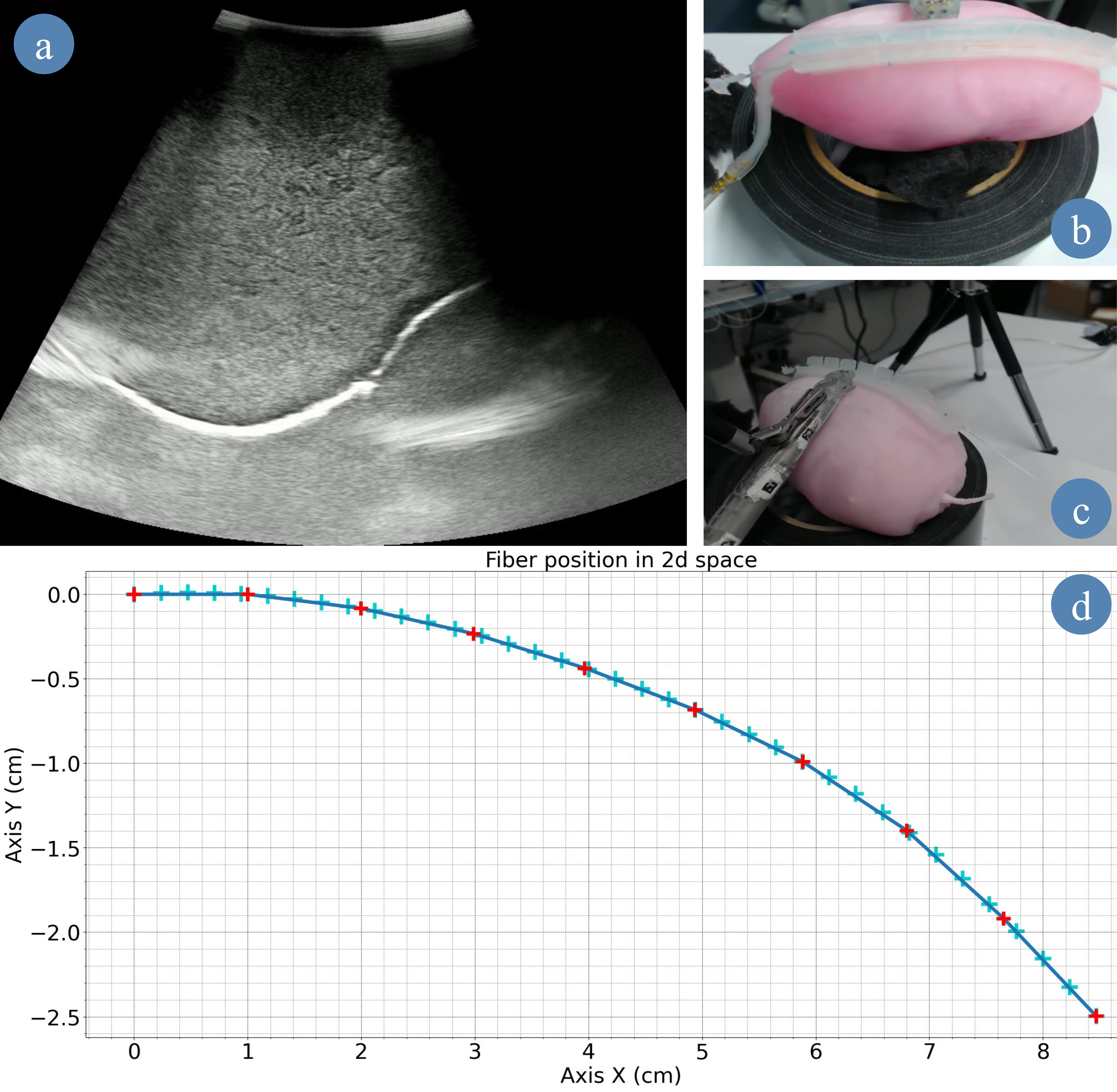}
    \caption{Case study snapshot. a) Stream snapshot recorded with the drop-in US probe. b) Camera top view. c) Camera lateral view. d) Computed drop-in US trajectory.}
    \label{fig:case_study_with_ultrasound}
\end{figure}

\subsection{Material Stiffness Experiments}
The experimental stress-strain curves were obtained for each of the tested materials. The Young's Modulus of the materials considered was obtained by applying a polynomial best-fit line to the data between 7.5$\%$ and 15$\%$ of maximum compression. We summarised these values in Table \ref{table:experimental_material_stiffness} alongside the documented Shore Hardness of the silicone materials.

\begin{table}
\centering
\caption{Experimental Material Stiffness Properties}
\label{table:experimental_material_stiffness}
\setlength{\tabcolsep}{25pt}
\begin{tabular}{ccc}
\hline
Material& 
Shore Hardness& 
Elastic Modulus (MPa)\\
\hline
Kidney Sample 1 & 
NA & 
($4.35 \pm 0.578) \times 10^{-3}$ \\
Kidney Sample 2 & 
NA &
($3.68 \pm 0.456) \times 10^{-3}$ \\
Kidney Sample 3 & 
NA& 
($5.79 \pm 0.672) \times 10^{-3}$ \\
\hline
%Eco-Flex\texttrademark 00-10&
%00-10&
%No data \\
Eco-Flex\texttrademark 00-20&
00-20&
$0.246 \pm 0.00230$ \\
Eco-Flex\texttrademark 00-30&
00-30&
$2.879 \pm 0.230$ \\
DragonSkin\texttrademark FX-Pro&
2 A&
$3.632 \pm 0.181$ \\
DragonSkin\texttrademark 10 NV&
10 A &
$7.660 \pm 0.211$ \\
DragonSkin\texttrademark 20 &
20 A &
$8.481 \pm 0.194$ \\
DragonSkin\texttrademark 30 &
30 A &
$9.990 \pm 0.600$ \\
\hline
\end{tabular}
\end{table}

\subsection{Case Study}
We slid the US probe along the rail for 20 mm before it detached from it. However, the contact ensured a stable imaging of the phantom, including the embedded mass simulating a tumour embedded into it, as shown in Fig. \ref{fig:case_study_with_ultrasound}a). Then, the probe disconnected from the phantom and the US was lost. Multiple reasons can explain this trajectory error.

Since the current fibre integration design prevents the probe from sliding along the rail on one side, we attached the probe on the opposite side from the sensing fibre. In this experiment, we assumed that the shape of the kidney was uniform along the width of the rail. However, in reality, this is not the case. 
Additionally, we planned the trajectory of the US probe in a plane (as shown in  Fig.\ref{fig:case_study_with_ultrasound}d). In the 3D space, this can be sufficient control only if we ensure that the longitudinal plane of the rail is parallel to the trajectory plane. However, we manually positioned the probe perpendicular to the rail (as shown in Fig.\ref{fig:case_study_with_ultrasound}b-c), which might induce the trajectory error. With further development, we could better align these planes. For instance, we could use ArUco markers \cite{Romero-Ramirez2018SpeededMarkers} to compute the transformation matrix between them and update the probe's trajectory from the results.

Finally, the rail attachment only helps to slide the probe along the rail (See Fig.\ref{fig:case_study_with_ultrasound}b-c). The actual US element is located at 3 cm from this contact point. In our case, it helps us to obtain an US stream even if the probe attachment detaches from the rail. But, in further studies, we would need to integrate that information into the path planning of the probe.

\section{Conclusion}
This paper presents an integrated system for curvature sensing of the PAF rails achieved by integrating a multi-core FBG optical fibre into the body of the device. The system uses local strain measurements to sense curvature along the axis of the rails, therefore providing information on the local curvature of the rail without the need for additional sensors. The accuracy of the curvature sensing has been evaluated by comparing the sensed curvature to a range of known curvatures in the range of a human kidney. The system has shown promising results; the bare FBGS fibre can sense these curvatures with mean error ranging between 1.44 $\%$ and 2.72$\%$ over the range of radii (when considering the central subsection of the sensing portion of the fibre). 

The accuracy of the curvature sensing when the fibre is embedded in the PAF rail of different materials has also been evaluated. The DS30 rail was identified as the optimal rail material based on sensing accuracy, reliability and material properties. When this material is used, the FBG sensing system can sense curvatures between $30$ mm and $110$ mm with a mean error ranging between $1.35$\% at $110$ mm and $18.9$\% at $30$ mm.  Significant improvements can be made by further evaluating the systematic errors affecting the system when the rail is vacuumised and increasing the iterations of experiments from 5 to 10 in order to reduce the variance.

Furthermore, we have demonstrated the ability to use the sensed curvature to control the trajectory of the da Vinci surgical robot in real-time. Using the da Vinci Research Kit, we autonomously planned and executed the trajectory of the US probe paired with the PAF rail during a kidney phantom scan (single swipe). We inferred the trajectory from the sensed curvatures of the FBGS system. Qualitatively the resultant images compared well with those achieved by manual execution. We propose this system for autonomous or semi-autonomous guidance of the US probe by the da Vinci surgical robot to achieve a more stable scan and improved ultrasound image than manual swiping. 

Future work will focus on quantitatively evaluating the effectiveness of the FBGS-sensed curvature to plan the path of the drop-in ultrasound probe on the surface of the kidney, understanding the systematic and random errors present in the curvature sensing performance of the sensorised PAF rails, and their impact in the trajectory executed by the da Vinci surgical robot. We will also look to involve clinicians in quantitative evaluation of the autonomously acquired US images compared with manual acquired ones.

% \section{Nomenclature}

% \subsection{Resource Identification Initiative}
% To take part in the Resource Identification Initiative, please use the corresponding catalog number and RRID in your current manuscript. For more information about the project and for steps on how to search for an RRID, please click \href{http://www.frontiersin.org/files/pdf/letter_to_author.pdf}{here}.

% \subsection{Life Science Identifiers}
% Life Science Identifiers (LSIDs) for ZOOBANK registered names or nomenclatural acts should be listed in the manuscript before the keywords. For more information on LSIDs please see \href{https://www.frontiersin.org/about/author-guidelines#Nomenclature}{Inclusion of Zoological Nomenclature} section of the guidelines.

% \section{Additional Requirements}

% For additional requirements for specific article types and further information please refer to \href{http://www.frontiersin.org/about/AuthorGuidelines#AdditionalRequirements}{Author Guidelines}.

\section*{Conflict of Interest Statement}
%All financial, commercial or other relationships that might be perceived by the academic community as representing a potential conflict of interest must be disclosed. If no such relationship exists, authors will be asked to confirm the following statement: 

The authors declare that the research was conducted in the absence of any commercial or financial relationships that could be construed as a potential conflict of interest.

\section*{Author Contributions}
A.S, and E.D. conceived the presented idea. A.M. developed the prototypes and experimental hardware. S.D. developed the software. A.M. and S.D. designed and carried out the experiments, and performed the analysis. A.M., S.D., and J.C. contributed to interpretation of results. A.M. wrote the manuscript with input from S.D., J.C., E.D., L.L., A.S. and D.S.. A.S. and D.S. supervised the project.

\section*{Funding}
This research was funded in whole, or in part, by the Wellcome/EPSRC Centre for Interventional and Surgical Sciences (WEISS) [203145/Z/16/Z]; the Engineering and Physical Sciences Research Council (EPSRC) [EP/P012841/1]; and the Royal Academy of Engineering Chair in Emerging Technologies Scheme [CiET1819/2/36]. For the purpose of open access, the authors have applied a CC BY public copyright licence to any author accepted manuscript version arising from this submission.

% \section*{Acknowledgments}
% This is a short text to acknowledge the contributions of specific colleagues, institutions, or agencies that aided the efforts of the authors.

% \section*{Supplemental Data}
%  \href{http://home.frontiersin.org/about/author-guidelines#SupplementaryMaterial}{Supplementary Material} should be uploaded separately on submission, if there are Supplementary Figures, please include the caption in the same file as the figure. LaTeX Supplementary Material templates can be found in the Frontiers LaTeX folder.

% \section*{Data Availability Statement}
% The datasets [GENERATED/ANALYZED] for this study can be found in the [NAME OF REPOSITORY] [LINK].
% Please see the availability of data guidelines for more information, at https://www.frontiersin.org/about/author-guidelines#AvailabilityofData

\bibliographystyle{Frontiers-Harvard} %  Many Frontiers journals use the Harvard referencing system (Author-date), to find the style and resources for the journal you are submitting to: https://zendesk.frontiersin.org/hc/en-us/articles/360017860337-Frontiers-Reference-Styles-by-Journal. For Humanities and Social Sciences articles please include page numbers in the in-text citations 
\bibliography{frontiers}

\end{document}